# Detecting socially interacting groups using f-formation: A survey of taxonomy, methods, datasets, applications, challenges, and future research directions

HRISHAV BAKUL BARUA, Robotics & Autonomous Systems, TCS Research, India
THEINT HAYTHI MG, Myanmar Institute of Information Technology, Myanmar
PRADIP PRAMANICK, Robotics & Autonomous Systems, TCS Research, India
CHAYAN SARKAR, Robotics & Autonomous Systems, TCS Research, India

Robots in our daily surroundings are increasing day by day. Their usability and acceptability largely depend on their explicit and implicit interaction capability with fellow human beings. As a result, social behavior is one of the most sought-after qualities that a robot can possess. However, there is no specific aspect and/or feature that defines socially acceptable behavior and it largely depends on the situation, application, and society. In this article, we investigate one such social behavior for collocated robots. Imagine a group of people is interacting with each other and we want to join the group. We as human beings do it in a socially acceptable manner, i.e., within the group, we do position ourselves in such a way that we can participate in the group activity without disturbing/obstructing anybody. To possess such a quality, first, a robot needs to determine the formation of the group and then determine a position for itself, which we humans do implicitly. The theory of f-formation can be utilized for this purpose. As the types of formations can be very diverse, detecting the social groups is not a trivial task. In this article, we provide a comprehensive survey of the existing work on social interaction and group detection using f-formation for robotics and other applications. We also put forward a novel holistic survey framework combining all the possible concerns and modules relevant to this problem. We define taxonomies based on methods, camera views, datasets, detection capabilities and scale, evaluation approaches, and application areas. We discuss certain open challenges and limitations in current literature along with possible future research directions based on this framework. In particular, we discuss the existing methods/techniques and their relative merits and demerits, applications, and provide a set of unsolved but relevant problems in this domain.

Authors' addresses: Hrishav Bakul Barua, Robotics & Autonomous Systems, TCS Research, Kolkata, India, hrishav.barua@tcs.com; Theint Haythi Mg, Myanmar Institute of Information Technology, Mandalay, Myanmar, theinthaythimg@gmail.com; Pradip Pramanick, Robotics & Autonomous Systems, TCS Research, Kolkata, India, pradip.pramanick@tcs.com; Chayan Sarkar, Robotics & Autonomous Systems, TCS Research, Kolkata, India, sarkar.chayan@tcs.com.

## 1 INTRODUCTION

Human group [116] and activity detection [1] has been a hot topic for computer/machine vision (CV/MV), Artificial Intelligence (AI), and robotics research. When humans interact with each other in a group (two or more people), they use common sense to position themselves concerning each other which facilitates easy interaction in the situation. The same thing is also applied when a new person wants to join a group. That person also assesses (implicitly) which place is best for joining so that people in the group do not face any inconvenience. The person also considers her role in that group according to the organization they are in right now to position herself [40]. Nowadays, robots are widely used in our daily surroundings for many purposes. One such popular application is to use a robot to attend meetings/conferences/discussions remotely as a telepresence medium. In such scenarios, the robot has to join a group of people. For the robot to fluidly participate in groups, they must need to know how the groups are formed, how they are shaped, and how they have evolved [57]. There can be many kinds of groups that defer in dimension, situation, organization, etc., and they are generally referred to as “f-formation”.

F-formation (facing formation) is defined as the set of patterns that are formed during social interactions of two or more people. A robot can join an existing group or go to a single person and form a new group [42, 46]. There are three social spaces related to f-formation, which are: O-space, P-space, and R-space. O-space is known as the joint transaction space which is the interaction space between participants. P-space is the space where active participants are standing. R-space is the area that surrounds the participants and is outside the interaction radius as shown in Fig. 1 (details in Section 2). According to social science, Kendon (1990) proposed four formations as standard formations. They are vis-a-vis/face-to-face, side-by-side, L-shaped, and circular. Apart from these, there are also many other kinds of formations such as semi-circular, rectangle, triangular, v-shaped, and spooning [125], [72]. By categorizing a formation to any of these formation types, a robot can understand how people are standing in the discussion, and decide a position for itself to join the group accordingly. In joining, the robot should take care of the fact that already standing people are not disturbed and obstructed by itself. In detecting groups, several methods exist such as determining the position and orientation of people, Graph-cuts methods, Hough-voting system, etc. One major problem in f-formation detection is occlusion [41]. People in a group may stand in such positions that some of them may occlude the others, or the viewing camera angle is placed in such a location that the complete group is not visible. In such cases, a robot will not know the formation as it may not be able to detect the bodies of some people. So, in this kind of situation, we have to think about what kind of formation we will assume for the robot to continue its work.

***Motivation and Research Objective.*** Social group and interaction detection is a non-trivial task in computer vision and is of paramount importance to the robotics research community. Many research groups across the globe have concentrated their studies in this area. The idea of social groups and interactions was first proposed in 1990 by Kendon [70]. However, the first mention of human proxemics dates back to 1963 [55]. But, back then no one knew that after a few decades these concepts of f-formation and proxemic behaviors of humans will be a basis for interaction detection in a group of people or robots or both. The research has gained pace since 2010 (see table 3), delivering many rule-based and learning-based methods and techniques for detecting interaction and f-formation in social setups for various applications, mostly robotics and vision. But not many surveys, reviews, or tutorials have been found in this domain which provides a good overall impression of the research, the state of the art, and future opportunities. This survey article is a 360° view into the problem of social group and interaction detection using f-formation covering almost all the concerns and aspects comprehensively. The aim is to demystify the domain and facilitate the scientists, researchers, and computer engineers to get a fair idea of the area and conduct fruitful research in the future. We also put forward a few related surveys in Table 1 and have compared them with our work based on various concern areas of our survey framework (discussed in Section 3 and Fig. 4).

***Uniqueness of the survey.*** To the best of our knowledge, this survey is the first of its kind in this subject area. Our survey puts forward the idea of social groups with the perspective of f-formation with some comprehensive details. We also discuss the optimum joining position for a robot to enable human-robot interaction, after successful detection of the formation using computer vision techniques. Additionally, we propose a holistic framework to signify the various concern areas in the detection and prediction of social groups. Various taxonomies, regarding camera view of the environment for collecting scenes for detection, datasets for training machine/deep learning (ML/DL) models, detection capability, and scale and evaluation methods are discussed. We discuss and categorize all the detection methods, particularly rule-based and machine learning-based. Furthermore, we also deliberate the application areas of such detection and recognition giving primary focus to robotics. We also detailed the challenges, limitations, and future research directions in this area.

***Organization of the survey.*** This survey article is organized into the following sections. Section 2 puts forward a comprehensive idea about social spaces involved in group interaction. The questions like the meaning of f-formation, types of f-formation, and evolution of f-formation from one type to another when a new member joins a group have been answered along with pictorial depiction for readers' understanding. In Section 3, we propose a generic and holistic framework for group and interaction detection using formations which also becomes the basis of categorization of the literature in the survey based on the various concern areas and modules. Then we present a year-wise compilation of research performed in this domain with analysis in Section 4. Section 5 discusses the various input methods for detection such as cameras and other sensors. It also puts focus on the various camera views and positions. Section 6 summarizes the methods, techniques, and algorithms for detection focusing on both rule-based static Artificial Intelligence (AI) methods and learning-based (data-driven) methods. Then we talk about detection capabilities and scale in Section 7. In this section, we also discuss briefly the various datasets available for training and testing purposes. Section 8 presents the various evaluation strategy and methodology from the perspective of algorithmic computational complexity and application areas like robotics and vision. The various application areas are stated in Section 9. Finally, we discuss the limitations and challenges in existing state-of-the-art literature and methods as well as propose some future research directions and prospects in each of the modules (of the survey framework) in Section 10. We conclude the survey in Section 11.

Table 1. Comparison of existing and related surveys/reviews with our work.

| **Existing Surveys and Reviews** | **(1)** | **(2)** | **(3)** | **(4)** | **(5)** | **(6)** | **(7)** | **(8)** | **(9)** | **(10)** | **(11)** |
|---|---|---|---|---|---|---|---|---|---|---|---|
| Tapus, Adriana, et al. [122] | ✗ | ✗ | ✗ | ✗ | ✗ | ✓ | ✗ | ✓ | ✓ | ✗ | ✗ |
| Setti, Francesco, et al. [116] | ✗ | ✗ | ✓✓ | ✓ | ✓ | ✓ | ✓ | ✓ | ✗ | ✗ | ✗ |
| Pathi, Sai Krishna, et al. [96] | ✗ | ✓ | ✗ | ✗ | ✓ | ✗ | ✓ | ✓ | ✓ | ✗ | ✗ |
| **This survey** | ✓✓ | ✓✓ | ✓✓ | ✓✓ | ✓✓ | ✓✓ | ✓✓ | ✓✓ | ✓✓ | ✓✓ | ✓✓ |

✗-> signifies no treatment in the paper, ✓-> signifies some mention exists and ✓✓-> means comprehensive treatment of the concern area. (1) Comprehensive F-formation list and tutorial on social spaces, (2) Camera views and sensors, (3) Datasets, (4) Detection capability/scale, (5) Evaluation methodology, (6) Feature selection, (7) Rule-based AI methods/techniques, (8) Machine Learning based AI methods/techniques, (9) Applications, (10) Limitations, Challenges and Future directions, (11) Generic survey framework for group and interaction detection.

## 2 SOCIAL SPACES IN GROUP INTERACTION

In this section, we describe the theory of f-formation, leveraging the theory to study the groups of interacting people, and how a robot can utilize this to imitate social behavior while interaction with a group of people.

### 2.1 What is f-formation?

Facing formation (f-formation) happens when two or more people sustain a spatial and orientational relationship and they have equal, direct, and exclusive access to the space between them [61]. Fig. 1 depicts such a social space where a group of people are interacting. An f-formation is the proper organization of three social spaces: O-space, P-space, and R-space [116]. They are situated like three circles surrounding each other. O-space, the innermost circle, is a convex empty space that is normally surrounded by the people in the group and the participants generally look inward into the O-space. P-space, the second circle, is a narrow space where active participants are standing. The R-space, the outermost circle, is a space where an inactive participant (listener) or an outsider who is not a part of the conversation stands.

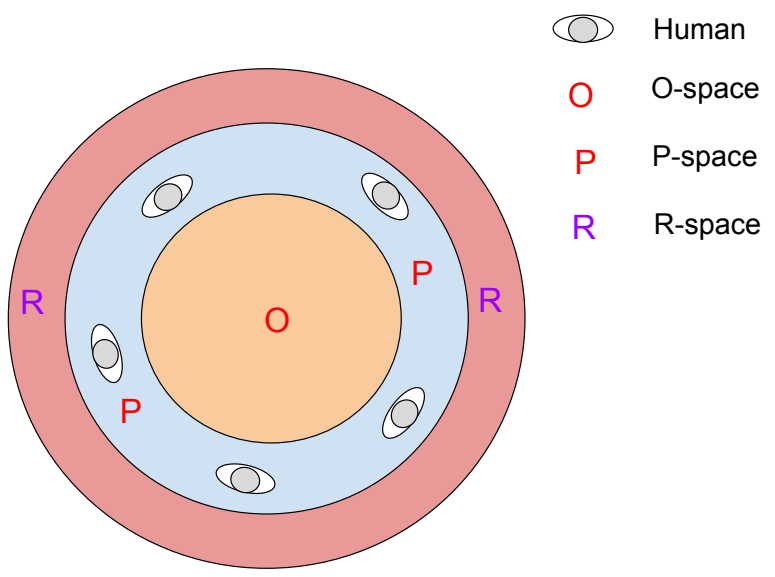

Fig. 1. The three social spaces corresponding to human proxemics.

### 2.2 Different f-formations possible during interaction

Although both the theory of f-formation and appropriate methods to detect them have been well-analyzed in the literature, a comprehensive list of all the possible f-formations during different kinds of interactions is yet to be brought out. The most common ones are *side-by-side, viz-a-viz, L-shaped and triangular*, defined for groups of two to three persons. Some others include *circular, square, rectangular and semi-circular*, which are more flexible and can contain a varying number of persons. We list down and categorize a complete collection of all the known f-formations below (see Fig. 2 for a pictorial representation).

(a) Side-by-side: The side-by-side formation is formed when two people stand close to each other facing in the same direction. Both faces either right or left or center. A minimum of 2 people is required for such a formation [61].
(b) Vis-a-vis or face-to-face: This formation comes into existence when two people are facing each other. Only 2 people are required for such a formation [61].
(c) L-shape: The L-shape is formed when two people face each other perpendicularly and are situated on the two ends of the letter "L" – one person facing the center and the other facing right or left [61].
(d) Reserved L-shaped: This is formed when two people are in a position of L-shape, but they are facing in different directions [72].
(e) Wide V-shaped. Two people are facing in the same direction like side by side but they tilt their bodies slightly to face each other a little. Minimum 2 people required for this formation [72].
(f) Spooning: This formation has two people with one person facing forward and the other look over from the back in the same direction [72].
(g) Z-shaped: This is formed when two people are standing side-by-side but facing in opposite directions [72].

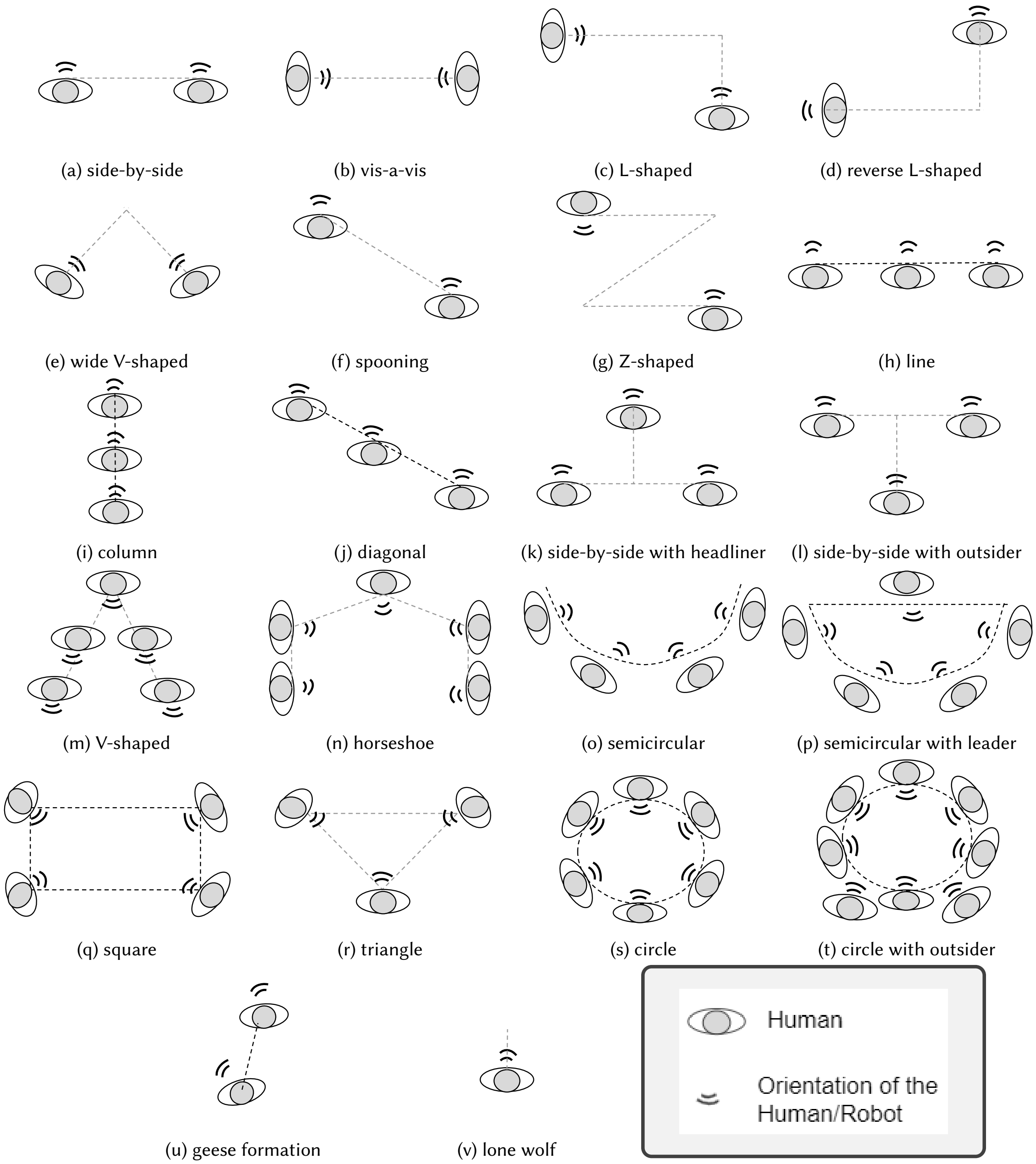

(a) side-by-side (b) vis-a-vis (c) L-shaped (d) reverse L-shaped
(e) wide V-shaped (f) spooning (g) Z-shaped (h) line
(i) column (j) diagonal (k) side-by-side with headliner (l) side-by-side with outsider
(m) V-shaped (n) horseshoe (o) semicircular (p) semicircular with leader
(q) square (r) triangle (s) circle (t) circle with outsider
(u) geese formation (v) lone wolf

Fig. 2. Comprehensive list of f-formations which are/can be used for group and interaction detection tasks in vision and robotics.

(h) Line formation: In this formation, all are standing in side-by-side fashion as a straight line and a minimum of 2 people are required [135].
(i) Column formation: In this formation, all are standing in a fashion where one is behind the other in a straight line and a minimum of 2 people are required [136].

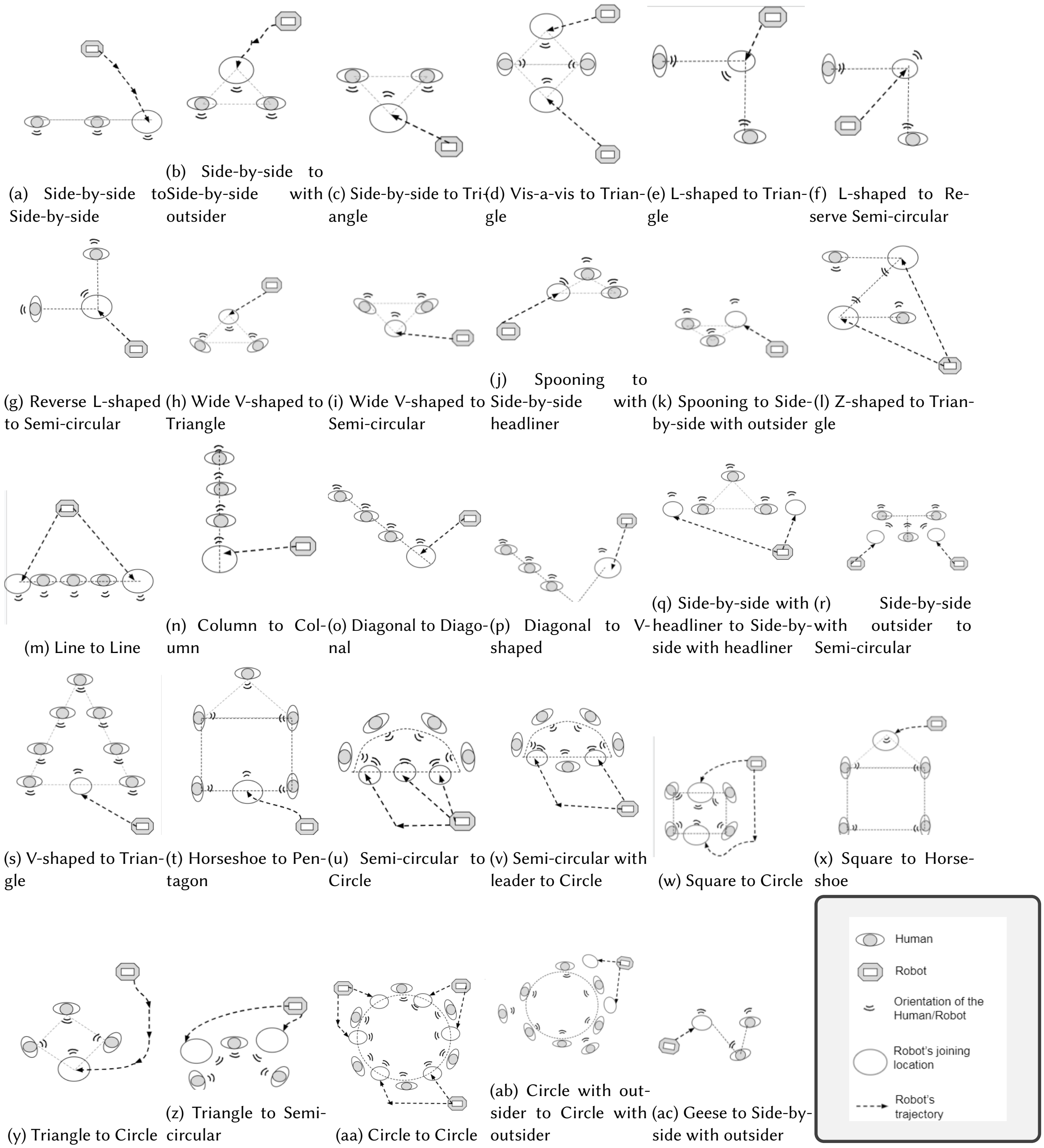

(a) Side-by-side to Side-by-side (b) Side-by-side to Side-by-side with outsider (c) Side-by-side to Triangle (d) Vis-a-vis to Triangle (e) L-shaped to Triangle (f) L-shaped to Reserve Semi-circular

(g) Reverse L-shaped to Semi-circular (h) Wide V-shaped to Triangle (i) Wide V-shaped to Semi-circular (j) Spooning to Side-by-side with headliner (k) Spooning to Side-by-side with outsider (l) Z-shaped to Triangle

(m) Line to Line (n) Column to Column (o) Diagonal to Diagonal (p) Diagonal to V-shaped (q) Side-by-side with headliner to Side-by-side with headliner (r) Side-by-side with outsider to Semi-circular

(s) V-shaped to Triangle (t) Horseshoe to Pentagon (u) Semi-circular to Circle (v) Semi-circular with leader to Circle (w) Square to Circle (x) Square to Horseshoe

(y) Triangle to Circle (z) Triangle to Semi-circular (aa) Circle to Circle (ab) Circle with outsider to Circle with outsider (ac) Geese to Side-by-side with outsider

Fig. 3. Comprehensive list of f-formations and their corresponding changed formations after a person/robot joins it. Also, the relevant possible trajectories a robot/person should take in each of the case.

Table 2. Comprehensive list of formations before and after a robot/human has joined. The "*" signifies that the number is the minimum requirement for that particular formation.

| No. | Formation (before) | No. of people | Formation (After joining) |
|---|---|---|---|
| a | Side-by-Side | 2* | Side-by-side, Side-by-side with one outsider, Triangle |
| b | Vis-a-vis | 2 | Triangle |
| c | L-shaped | 2 | Triangle, Reverse Semi-circular |
| d | Reserved L-shaped | 2 | Semi-circular |
| e | Wide V shaped | 2* | Triangle, Semi-circular |
| f | Spooning | 2 | Side-by-side with headliner, Side-by-side with outsider |
| g | Z-shaped | 2 | Triangle |
| h | Line | 3 | Line |
| i | Column | 3 | Column |
| j | Diagonal | 3 | Diagonal, V-shaped |
| k | S-by-s with one headliner | 3* | S-by-s with one headliner |
| l | S-by-s with outsider | 3* | Semi-circular |
| m | V-shaped | 7 | Triangle |
| n | Horseshoe | 5* | Pentagon |
| o | Semi circular | 4 | Circular |
| p | Semi-circular with one leader in the middle | 5 | Circle |
| q | Square | 4 | Circular, Horseshoe |
| r | Triangle | 3 | Circle, Semi-circular |
| s | Circle | 6 | Circle |
| t | Circle with outsider | 8 | Circle with outsider |
| u | Geese | 2* | S-by-s with outsider |
| v | Lone wolf | 1 | Vis-a-vis, S-by-s, L-shaped |

(j) Diagonal (Echelon): In this formation, people stand diagonally and face in the same direction. A minimum of 2 people is required [137].

(k) Side-by-side with one headliner: In this formation, one person stands in the front and others stand side-by-side at the back. A minimum of 3 people are required and they all face in the same direction [40].

(l) Side-by-side with outsider: In this formation, one participant occupying an outer position of the side-by-side formation in the r-space, who usually does not play an active role in the conversation. A minimum of 3 people is required [79].

(m) V-shaped: In this formation, all people stand in a V-shaped fashion and face the same direction. A minimum of 3 people is required [138].

(n) Horseshoe: The group of people stands in the shape of "U" and a minimum of 5 people are required [116].

(o) Semi-circular: The semi-circular formation is where three or more people are focusing on the same task while interacting with each other [96].

(p) semi-circular with one leader in the middle: In this formation, people stand in semi-circular shape and there is one person in the center who is facing to the group of people in the semi-circle. A minimum of 4 people is required [79].

(q) Square (Infantry square): Four people stand in the square shaped fashion [139].

(r) Triangle: As the name suggests, three people stands in a triangular shape in this formation [96].

(s) Circle: As the name suggests, a group of people stands in a circular shape in this formation [94].

(t) Circular arrangement with outsiders: In this formation, some people stand in a circular fashion and one/two additional people stand at the back of the circular formation [40].

(u) Geese formation: In this formation, there are two or more people where one person is leading the path and the others are following that person but may or may not be looking in the same direction. A minimum of 2 people is required [42].

(v) Lone wolves: This is not really a formation (yet). There is only one person ready to be joined by others before an interaction [42].

### 2.3 Best position for a robot to join in a group

As social robotics is one of the most important application areas of group and interaction detection using computer vision, we put forward a list of positions where a robot can join a formation (here f-formation) after successfully detecting it. But joining a group requires a socially aware [28] or human-aware [74] navigation protocol embedded into the robot. In other words, the robot should imitate human-like natural behavior while approaching a group (considering correct direction and angle) for interaction and discussion without incurring any discomfort to the existing members of the group. However, this part of the story is out of the scope of our survey and we limit our work to detection and prediction of group and interaction only. But, as it seems necessary to at least briefly mention this side of the coin, we put forward the possible joining locations and natural joining path or approach direction/angle (robot trajectory) in this section (also discussed briefly in Sections 8 and 10). Researchers can think of presenting and publishing a systematic survey on the navigation and joining aspect of a robot/autonomous agent after successful detection/prediction of the group interaction and f-formation.

Table 2 summarizes a list of formations with the number of people and correspondingly the new formations after a person/robot joins it. The pictorial summary of the same is presented in Fig. 3.

## 3 GENERIC SURVEY FRAMEWORK WITH POSSIBLE CONCERN AREAS

This survey aims to facilitate the concerned researchers with a comprehensive overview of this domain of group and interaction detection. The idea of group/interaction detection is not new and is around for more than a decade. Researchers are trying to design and develop new methods, techniques, algorithms, and architectures for various application areas ranging from computer vision and robotics to social environment analysis.

The problem of a group and/or interaction detection is a non-trivial problem of computer vision. The existing research approaches follow both classical AI algorithms like rule-based methods, geometric reasoning, etc., and neural networked-based methods. Moreover, learning paradigms like supervised, semi-supervised and unsupervised are also used. Proper categorization of these methods is necessary for future research directions. We have proposed a holistic framework that corresponds to the concern areas of f-formation research and can also be considered as a generic architecture for a typical group/interection detection task using f-formation. Fig. 4 puts forward a possible framework with different modules of such a detection task. The various concern areas of this domain can be characterized by −sensors used, camera view/position for capturing the group interaction, datasets used for training/testing the method in case of learning-based approaches (indoor or outdoor), feature selection method & criteria, detection capabilities (static/dynamic scenes) & scale (single or multi-group scenario), evaluation methodology (efficiency/accuracy and/or simulation study and human experience study), and application areas. The mentioned modules are used as the basis of categorization of the literature in our survey and are attended to in the upcoming sections one by one (as mentioned in Fig. 4). Finally, we conclude the survey by discussing the limitations, challenges, and future directions/prospects (Section 10) in each of the concerned modules.

## 4 RESEARCH CHRONOLOGY

In 1990, Kendon [70] proposed the f-formation theory for group interaction by participating people on the basis of proxemics behavior. A computer system to detect human proxemics behavior was first studied by Hall almost six decades ago [55, 56]. This section is a survey on the literature collected on f-formation, using static and learning-based AI approaches. Fig. 5 shows the various specified distance ranges for different designated interaction types on the basis of intimacy level between the participating people. The distance ranges specified in green colored boxes are relevant to group/interaction and f-formation detection perspective. The blue-colored boxes signify distance ranges that are not generally seen in any f-formation. After the early studies by Hall, it was until 2009 when computer vision-based approaches using static AI methods or ML/DL techniques have emerged. From 2010, the research in this field started gaining pace. From 2013 to 2017, the research in this domain reached its peak with multiple methods, algorithms, and techniques being proposed. From 2010 to 2013, rule-based fixed AI methods were prevalent. Learning-based or data-driven methods like DL and RL have gained prominence in 2014, and gaining pace with each passing year. These methods have attained popularity in 2019. The years from 2014 to 2018 have seen equal contributions in both fixed rule-based methods and learning-based methods. So, there is a transition from traditional AI-based methods to machine learning and data-driven techniques like almost any domain of AI. Table 3 can be referred to for the complete list of references (year-wise). The table also consists of the keywords (methods, focus areas, and technologies) for most of the references for a better perception of the readers.

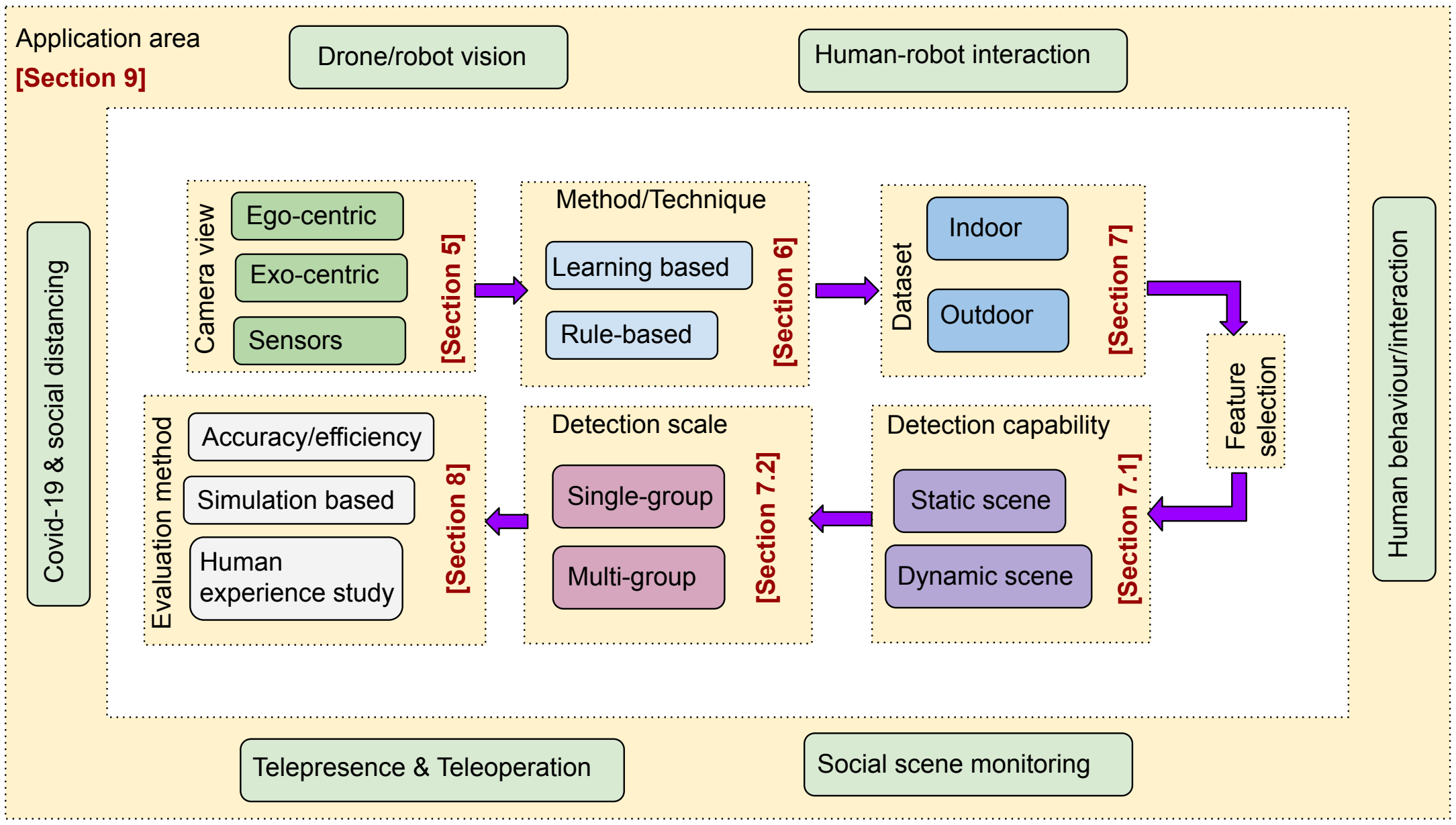

Fig. 4. A possible separation of concerns into modules regarding social groups/interactions detection. The arrows correspond to the flow of events/data in a typical group/interaction detection (f-formation) framework.

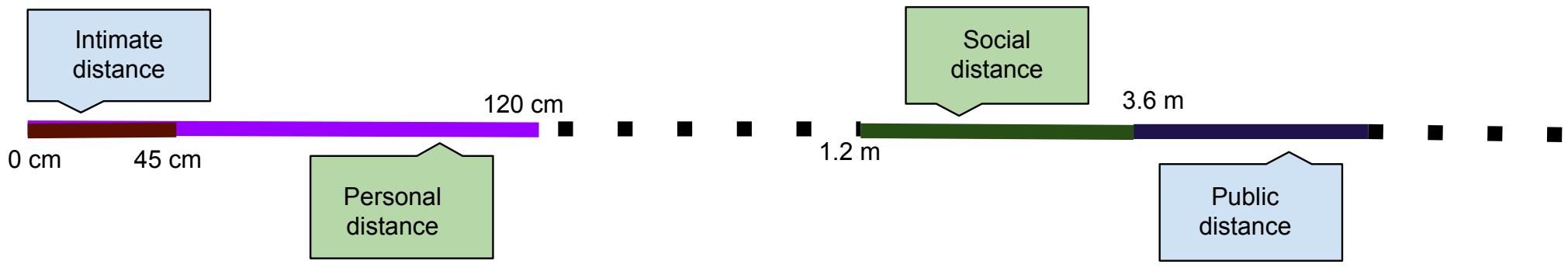

Fig. 5. Human proxemics using distance ranges as stated by Hall [55]. The green comment boxes signify distance ranges typically used for f-formation and group interaction.

Table 3. Year wise compilation of methods and techniques for group/interaction/f-formation detection and other relevant literatures.

| Year of Publication | Study, Methods and Techniques | Total |
|---|---|---|
| 2004 | Geometric reasoning on sonar data [12]. | 1 |
| 2006 | Wizard-of-Oz (WoZ) study on spatial distances related to f-formations [61]. | 1 |
| 2009 | Clustering trajectories tracked by static laser range finders [67], trajectory classification by SVM [112]. | 2 |
| 2010 | Probabilistic generative model on IR tracking data [54], WoZ study of robot's body movement [75], SVM classification using kinematic features [144]. | 3 |
| 2011 | Analysis of different f-formations for information seeking [79], Hough-transform based voting [37], graph clustering [60], a study on transitions between f-formations on interaction cues [82], a computational model of interaction space for virtual humans extending f-formation theory [88], a study of physical distancing from a robot [85], utilizing geometric properties of a simulated environment [86], a study to relate f-formations with conversation initiation [117], Gaussian clustering on camera-tracked trajectories [38]. | 9 |
| 2012 | Application of f-formations in collaborative cooking [91], Kinect-based tracking with rules [78], WoZ study on social interaction in nursing facilities [71], a study of robot gaze behaviors in group conversations [143], velocity models (while walking) [84], SVM with motion features [108], Hidden Markov Model (HMM) [49]. | 7 |
| 2013 | Spatial geometric analysis on Kinect data [44, 46], analysis of f-formation in blended reality [41], a comparison of [37] and [60] [114], exemplar based approach [77], multi-scale detection [115], Bag-of-Visual-Words (BoVW) based classifier [126], Inter-Relation Pattern Matrix [23], HMM classifiers [81], O-space based path planning [52]. | 10 |
| 2014 | Hough Voting (HVFF), Graph-cuts (GCFF) [100], game theory based approach[128], correlation clustering algorithm [11], reasoning on proximity and visual orientation data [42], effects of cultural differences [65], HMM to classify accelerometer data [59], iterative augmentation algorithm [31], adaptive weights learning methods [102], estimating lower-body pose from head pose and facial orientation [142], search-based method [45], study on group-approaching behavior [69], spatial activity analysis in a multiplayer game [66]. | 12 |
| 2015 | Robust Tracking Algorithm using TLD [9], GCFF based approach [116], Correlation Clustering algorithm [10], multimodal data fusion [8], spatial analysis in collaborative cooking [90], GIZ (Group Interaction Zone) detection method [30], study on influencing formations by a tour guide robot [68], joint inference of pose and f-formations [121], participation state model [118], SALSA dataset for evaluating social behavior [7], multi-level tracking based algorithm [131], Structural SVM (SSVM) using Dynamic Time Warping (DTW) loss [119], Long-Short Term Memory (LSTM) network [2], influence of approach behavior on comfort [15]. | 14 |
| 2016 | F-formation applied to mobile collaborative activities [125], subjective annotations of f-formation [145], game-theoretic clustering [129], study of display angles in museum [62], mobile co-location analysis using f-formation [113], proxemics analysis algorithm [104], review of human group detection approaches [123], LSTM based detection in ego-view [3]. | 8 |
| 2017 | Haar cascade face detector based algorithm [73, 94], weakly-supervised learning [127], temporal segmentation of social activities [35], omnidirectional mobility in f-formations [141], review of multimodal social scene analysis [6], 3D group motion prediction from video [64], survey on social navigation of robots [28], a study on robot's approaching behavior [16], heuristic calculation of robot's stopping distance [109], a study on human perception of robot's gaze [130], computational models of spatial orientation in VR [97]. | 12 |
| 2018 | Optical-flow based algorithm in ego-view[105], meta-classifier learning using accelerometer data [50], human-friendly approach planner [111], discussion on improved teleoperation using f-formation [92], effect of spatial arrangement in conversation workload [80], study of f-formation dynamics in a vast area [40]. | 6 |
| 2019 | Study on teleoperators following f-formations [96], analysis on conversational unit prediction using f-formation [103], empirical comparison of data-driven approaches [57], LSTM networks applied on multimodal data [107], robot's optimal pose estimation in social groups [95], review of robot and human group interaction [140], Staged Social Behavior Learning (SSBL) [47], Euclidean distance based calculation after 2D pose estimation [93], Robot-Centric Group Estimation Model (RoboGEM) [124]. | 9 |
| 2020 | Difference in spatial group configurations between physically and virtually present agents [58], Conditional Random Field (CRF) with SVM for jointly detecting group membership, f-formation and approach angle [18]. | 2 |

## 5 CAMERAS AND SENSORS FOR SCENE CAPTURE

This section summarizes the input methods in the group/interaction detection framework (Fig. 4). The main input methods are cameras and sensors. There are different types of cameras used in the literature such as omnidirectional camera, helmet camera, robot camera, fisheye camera, and webcams. The camera sensor may be equipped with depth perception or provide only RGB images. The other main sensors found in the surveyed literature are audio sensors, blind sensors, and RFID (Radio-frequency identification) sensors, etc. These are chosen based on the application areas and the working environment.

### 5.1 Camera Views

There are two different types of camera positioning used – ego-vision/ego-view (ego-centric) camera for robotics and exo-vision/exo-view (exo-centric) or global view cameras (fixed in walls and ceiling) in indoor environments or outdoor environments (see Fig. 6). Cameras are used for drone surveillance, robotic vision, and scene monitoring. In these cases, we work with the ego/exo views of the scene to detect group interactions.

**Ego-centric view.** Ego-centric refers to the first-person perspective; for example, images or videos captured by a wearable camera or robot camera. The captured data is focused on the part where the target objects are placed. In [105], the robot's camera is used for capturing scenes which is also referred a robot-centric view. In [45], the authors use first-person view cameras for estimating the location and orientation of the people in a group. In [2], the authors use low temporal resolution wearable camera for capturing groups' images.

**Exo-centric view.** The exo-centric view is concerned with the third-person perspective or the top view; for example, images or videos which are captured by surveillance/monitoring cameras. There can be one or many social interaction groups in a scene that can be captured simultaneously from the top view. In [102], the authors use 4 cameras for detecting groups at large scale. The method also detects changes in the target groups when it moves closer or further from the cameras. In [60], experiments are done by capturing a video with a camera from approximately 15 meters overhead. In [38], the images are captured by using a fisheye camera and it is mounted 7 meters above the floor.

### 5.2 Other Sensors

Sensors play a vital role to find the relative distance of the people in a group, which helps accurate prediction of the type of f-formation. Researchers used different types of sensors such as depth sensor, laser sensor, audio or speech sensor, RFID, and UWB sensor in the literature. There are some cases where both cameras and other types of sensors are used simultaneously for detection. In [130] the authors use the UWB (Ultra-wideband) localization beacons, Kinect and an audio sensor for detecting people and other entities, and RGB cameras for monitoring. The data for scenes are captured in the form of images and/or videos depending on the method that uses the input for scene detection. Some instances of WiFi-based tracking [51] of humans are also visible in the literature.

Fig. 7 shows a taxonomy of cameras/vision sensors and other sensors used in the literature for scene capture. Table 4 gives a categorization of the surveyed literature on the basis of camera views and sensors. In the table, readers may see the number of cameras used in each of the cited papers specified. The table also specifies the various cameras and sensors used in each paper.

## 6 CATEGORIZATION OF METHODS/TECHNIQUES

There are many f-formation detection methods proposed in the literature. In this article, we broadly categorize these methods into two classes – (a) rule-based methods (fixed rules, assumptions, and geometric reasoning) like the conventional image processing and vision techniques, and (b) learning-based method (or data driven approach). With the Big data revolution at its bloom, learning-based methods have come to prominence in the recent past. Multimedia and visual analytics [99] from big data remains a lucrative tool for the future of this domain.

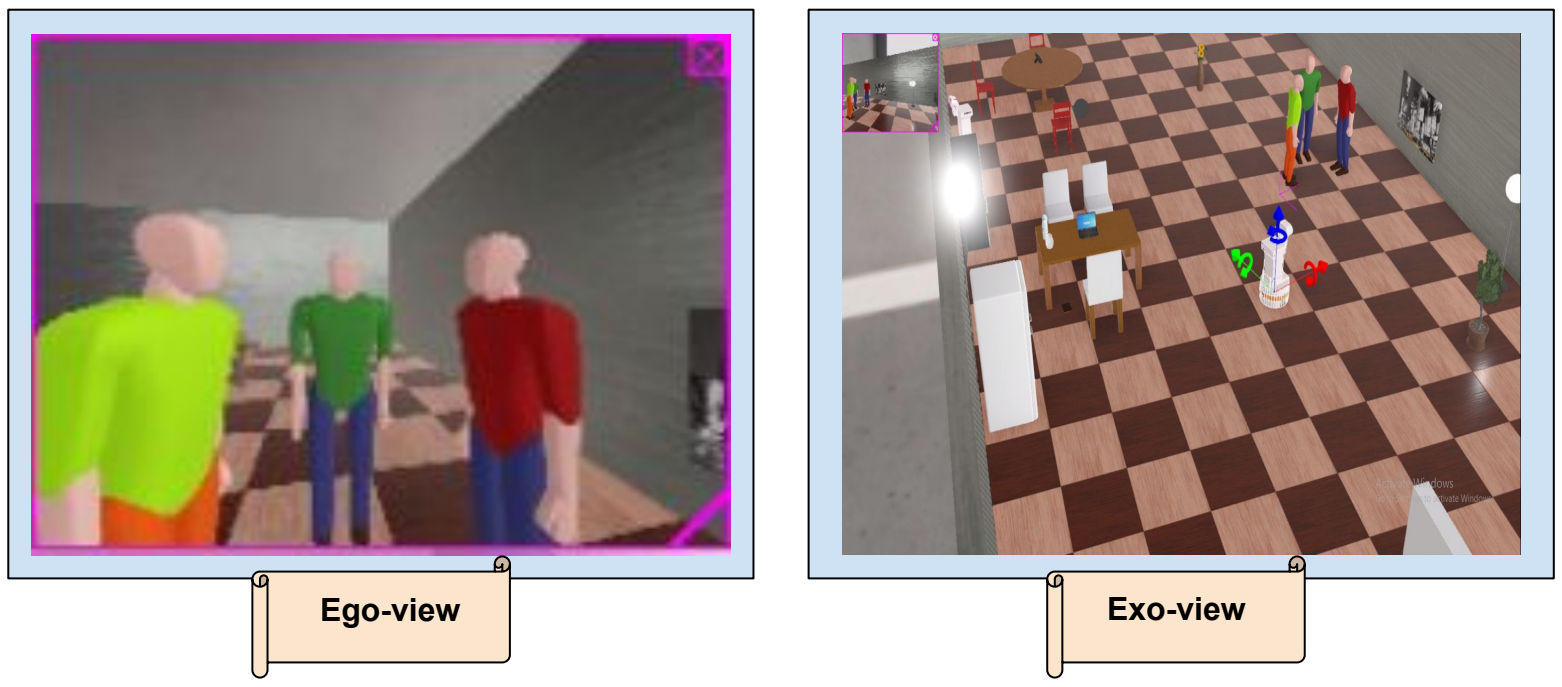

Fig. 6. Different camera views of the same group/interaction in an indoor environment. In the left side a robot is having a ego-view camera and in the right side an exo-view or global view camera is fixed in a wall. These images are produced in webot robotic simulator [39].

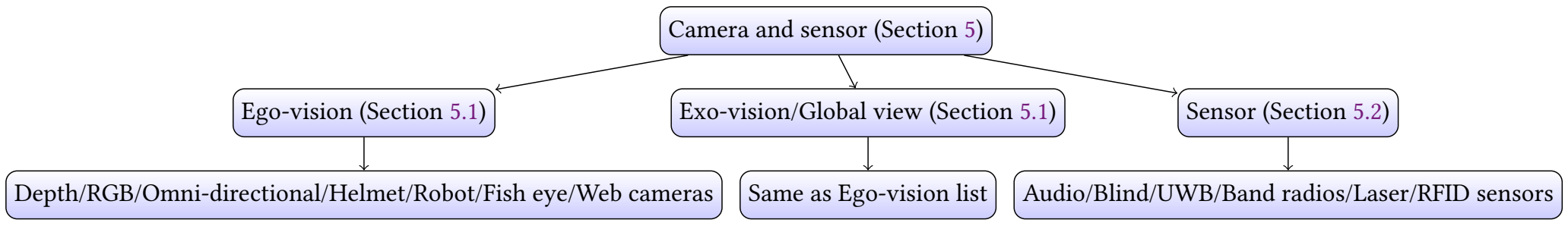

Fig. 7. Taxonomy for cameras and sensors for scene capture. The leaf nodes give examples in each category.

In group discussions, people stand in a position where the conversation can happen effectively. Kendon [70] proposed a formal structure of group proxemics among the interacting people in a formation (described in Section 2), where Hough-voting strategy is used for finding the O-space [100]. In [115], the authors use the Hough-voting approach with a two-step algorithm – 1) fixed cardinality group detection, and 2) groups merging. Using these two steps, they detect the type of f-formation. In [46], the experiment uses a heat map-based method for recognizing human activity and the best view camera selection method. In [100], the (GCFF) graph cut f-formation is used for detecting f-formations in static images with the graph-cuts algorithms via clustering graphs. Yasuharu Den [40] says that formations are also dependent on the social organization and environment. He explains formation with outsiders where people stand based on their position. In this paper[96], there are three constraint-based formations namely triangle, rectangle, and semi-circular formation. They use a game-theoretic model for the position and orientational information of people to detect groups in the scene. For checking the formation, they use an algorithm proposed by Vascon *et al.* [128, 129] that generates the 2D frustum of the position and orientation of people in the group. In [94], the authors use the Haar cascade face detector algorithm to detect the faces and eyes of people. Based on the face and eye detection, the method decides how many frontal, right, and/or left faces are there and then decides the formations. In [73], the Haar cascade classifier is used with quadrant methodology. The paper differentiates the person's facing direction by looking where the eye is located and in which quadrant. In [46], the authors use a new method to find the dominant sets and then compares with modularity cut. But this method is applicable only when everyone is standing. In [103], the method uses speaking turns for indicating the existence of distinct conversation floors and gets the estimation of the presence of voice. But this method cannot detect the silent (inactive) participants. In [107], proximity and acceleration data are used and pairwise representations are used with LSTM (Long short-term memory) network. They are used for identifying the presence of interaction and the roles of participants in the interaction. However, using a fixed threshold for identifying speakers can create mislabel in some instances. In [10], structural SVM

Table 4. Classification based on camera view and other sensors for group/interaction and f-formation detection.

| Classification | Application areas/ Details | References |
|---|---|---|
| Ego-centric (First person view or Robot view) [Section 5.1] | • Robotics and Human-robot interaction<br>• Robot vision in telepresence<br>• Drone/Robot surveillance | [58], 1[93],1[47], 1[124], 1[96], 2 Hamlet cameras and 1 robot camera[95], 1[111], 1[92], 1[105], 1[73], 1[94], multi [64], 1[3], 2[62], 1[123], 1[125], 1[2], 1[118], 1[68], 1[9], multi[90], multi[45], multi[142], 1[31], [69], multi[42], 1[11], 1[51], depth camera, RGB camera[49], an omni-directional camera[143], multi[91], [88], 3[75], 4[54], [57], robot camera[61], 2[12], [10], 2[67], 1[18] |
| Exo-centric (Global view) [Section 5.1] | • Social scene monitoring<br>• Covid-19 social distancing monitoring<br>• Human interaction detection and analysis | 1[96], multi[50], multi[92], 1[16],[141], multi[6], multi[64], 1[28], 4[130], multi[127], 8[109], 1[62], multi [123], multi[129], 4[104], 2[145], multi[119], [131], multi[7], multi[121], [116], 1[30], multi[116], multi[8], multi[90], [66], 1[48], 4[102], a single monocular camera[128], 3 overhead fish-eye camera used for training classifier [59], multi[142], multi [45], 1[31], 1[23], multi[44], 1[81], 4[115], [77], 1[126], multi[114], 7[46], 4 [71], 1[108], 2 [78], 1[29], multi [91], 1[38], 1[82], multi[37], 1[85], 1[60], multi[144], 4+2[54], 4 webcams [61], an omnidirectional camera [12], [100] , [146] |
| Using other Sensors [Section 5.2] | Audio, sociometric badges, Blind sensor, prime sensor, WiFi based tracking, laser based tracking, depth sensor , band radios, touch receptors, RFID sensors, smart phones, UWB becon | [7], [42], Kinect depth sensor [44], [51], [49], [82], [125], speakers[103], wearable sensors[107], [28], [109], ulta wide-band localization beacons ( UWB), Kinect [130], [67], [112], [113], RFID tag[62], [74], [44], [84], [88], Asus Xtion Pro sensor[93], ZED sensors [105], single worn accelerometer [50], Kinect sensor [111], Microsoft Speech SDK [97], speaker, Asus Xtion Pro live RGB-D sensor [16], Kinect [64], motion tracker[109], sociometric badges[6], RGB-D sensor [141], tablets [125], tablets [113], mobile sensors [123], microphone, infrared(IR) beam and detector, bluetooth detector, accelerometer [7], touch sensor [88], range sensor [143], laser sensors [84], Wi-fi based tracking, Laser-based tracking [51], PrimeSensor, Microsoft Kinect, microphone [81], RFID sensors [41], blind sensor, location beacon [42], single worn accelerometer [59], [80], gaze animation controller [97], [117], grid world environment [86], ethnography method [79] |
| Others relevant literatures | - | [55], [65], [52], [41], [70], [72], [40] |

(support vector machine) is used for learning how to treat the distance and pose information, and correlation clustering algorithm is used to predict group compositions. Furthermore, TLD (Tracking learning detection) tracker is used for blur detection for ego-vision images. But the trackers cannot perform detection when the target is moving out of the camera field of view. In [105], the method uses ego-centric pedestrian detection. The pedestrian detector generates bounding boxes. It uses optical flow for estimating motion between consecutive image frames. For detecting groups, they used joint pedestrian proximity and motion estimation. In [145], the method detects the group with a group detector first then uses the trained classifier to differentiate the people involved in the group. Some researchers use pedestrians, vision-based algorithms, pose-estimation algorithm to detect [127] groups. In the article [127], authors use body-pose for handling f-formation detection and finding the joint estimation of f-formation and target's head and body orientation. They also use multiple occlusion-adaptive classifiers. There are many more methods scientists use but each of them has its own strengths and weaknesses. Fig. 8 presents a taxonomy of different methods/techniques/approaches surveyed in this article that are used for group/interaction/formation detection.

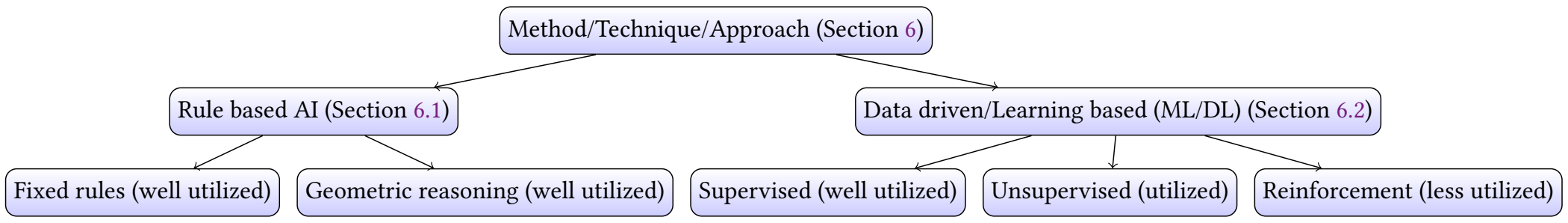

Fig. 8. Taxonomy for methods and approaches used for group and interaction detection.

### 6.1 Rule based method

We categorize methods as rule-based that include pre-defined rules, geometric assumptions, and reasoning. Rule-based methods are designed around well-known social behaviors and geometric properties and are often intuitive. In the absence of any learning paradigm, the algorithms are purely based on a static set of rules that are assumed to be true for a particular group situation (see Fig. 9).

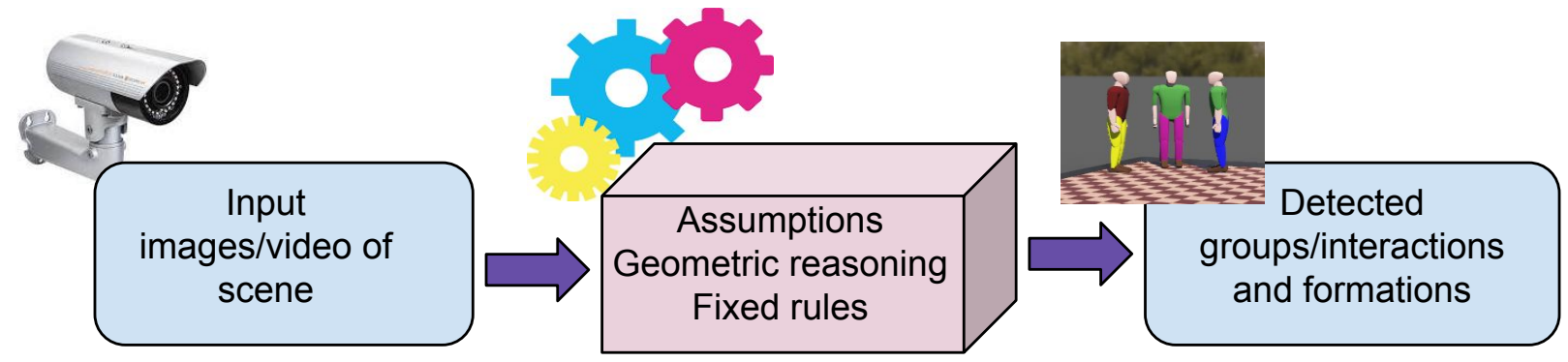

Fig. 9. Generic framework for a rule based AI method of detecting group interactions and formations.

In the following, we list down the most popular rule-based methods that report a decent accuracy in detecting human groups.

**Voting based approach (2013).** This approach is used for detecting and localizing groups by finding the matches based on exemplars. The authors in [77] suggest, this method works on agents so it is very flexible for different multi-agent scenarios. The results show that this method is effective for groups of up to four agents. The results are evaluated with people only without robots. The computational complexity of this method is low, hence it is real-time in nature and accuracy is very good.

**Graph cuts for f-formation (GCFF) (2015) [116].** GCFF approach firstly finds the o-space and gives the individual position to identify the orientational formation. This method is tested on a synthetic scenario and compared with other methods such as Inter-Relation Pattern Matrix (IRPM), Dominant Sets (DS), Interacting Group Discovery(IGD), Game-Theory for Conversational Groups (CTCG), Hough Voting for f-formation (HVFF), etc. This approach improves over other approaches not only in terms of precision but also in recall scores. It performs better in detecting people and orientation with no errors. The results are evaluated with people only.

**Head and body pose estimation (HBPE) (2015) [121].** This method uses a joint learning framework for estimating head and body orientations that in turn is used for estimating f-formations. This method is evaluated with people in a scene without any robots. For evaluating, authors use the mean angular error for head, body pose estimation (HBPE) and use F1-score for f-formation estimation (FFE). This method is compared with Hough Voting for f-formation (HVFF) method. Though the results are more or less similar, this method is slightly more accurate and has a higher F1 score.

**GROUP (2015) [131].** The GROUP algorithm detects f-formations based on lower-body orientation distributions of people from the scene and gives a set of free-standing conversational groups on each time step. Firstly, it analyzes the maximum description length (MDL) parameter. The higher the MDL parameter, the higher is the radius of grouping people together. This method can also detect non-interacting people as outliers. It is evaluated with people only without any robots in the scene. The computational complexity of this method has been compared with state-of-the-art methods.

**Approach Planner (2018) [111].** The Approach Planner (AP) enables a robot to navigates/plan based on the natural approaching behavior of humans toward the target person. This method can replicate human behavior/tendencies when approaching. The evaluation is based on the parameters derived from the skeletal information.

**Game-theoretic model (2019) [96].** The approach gives a 2D frustum for each virtual agent and robot by giving the position and orientation of them. Then it computes the affinity matrix. The method is evaluated both quantitatively and qualitatively. It is efficient in serving teleoperated robots who follow f-formations while joining groups automatically. The method also takes care of the fact that the formation is modified when new people/robots join the old group. The evaluation is made in a simulation environment.

### 6.2 Machine learning based method

Machine learning-based methods are generally data-driven models where different algorithms are explored by researchers. Generally, we have special case of Deep learning methods under Machine learning. The primary learning paradigms used are – Supervised, Unsupervised, Semi-supervised and Reinforcement learning. However, a generic system that uses a machine learning algorithm to detect f-formation is shown Fig. 10. In the following, we describe the various ML based approaches.

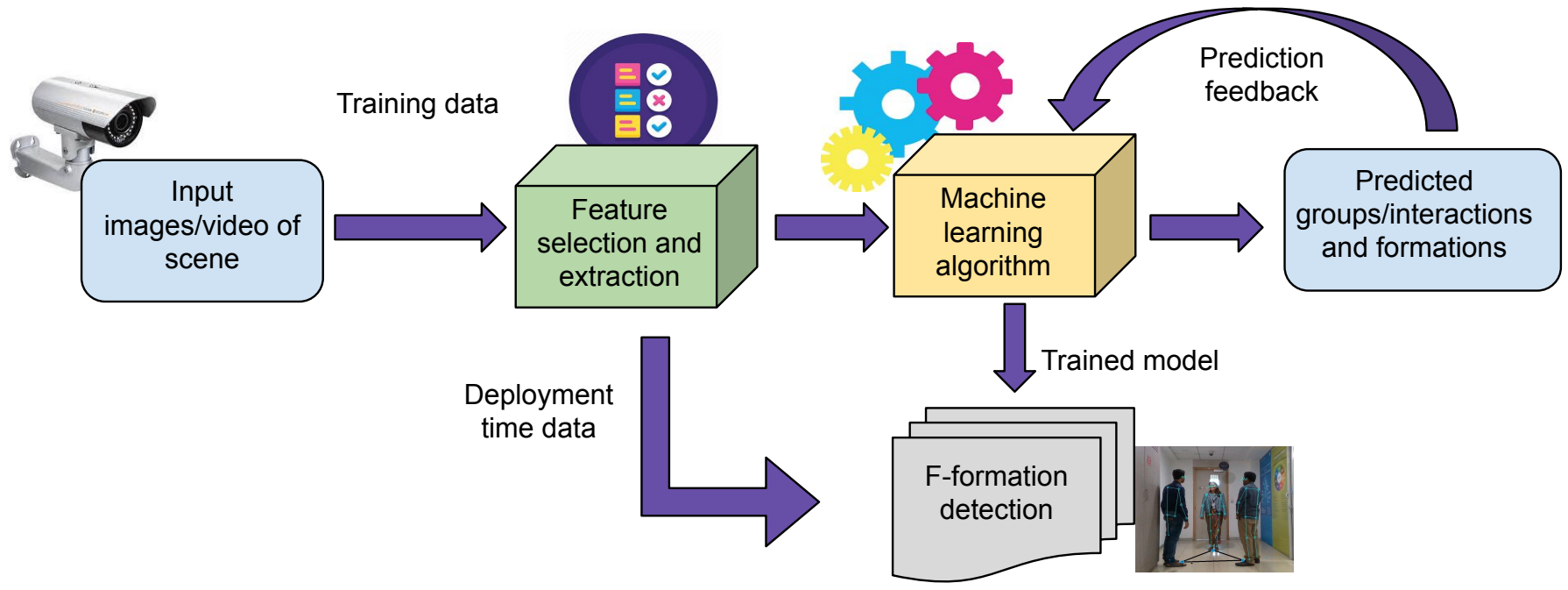

Fig. 10. Generic framework for a learning based AI method of detecting group interactions and formations.

**IR tracking method with SVM (2010) [54].** With the help of IR tracking, social interactions can be classified as either existing or non-existing by using geometric social signals. The authors train and test with many classifiers such as Support Vector Machines (SVM), Gaussian Mixture Models (GMM), and Naive Bayes classifier. IR tracking with an SVM classifier has been shown to achieve better accuracy than other classifiers.

**Graph-based clustering method (2011) [60], (2013) [126].** In [60], authors use the "socially motivated estimate of focus orientation" (SMEFO) feature to estimate body orientation that in turn estimates f-formation. This method has been compared with a modularity cut method. The evaluation process is done on the basis of computation complexity. The limitation of this approach can be seen in scenarios where people are moving within the group and/or people are joining/leaving the group.

In [126], authors find a graph representation from the 3D trajectories of people and head poses. Using a graph-clustering algorithm they discover social interaction groups. They use a Support Vector Machine (SVM) classifier for learning and classifying the group activities. The evaluation shows that it is better than the previous methods. Human experience study is also performed with robotic scenarios. This approach not only recognizes or detects particular group activity but also predicts a direct link between each person from that group.

**Novel framework (2013)[23].** This approach uses the Subjective View Frustum (SVF) as the main feature which encodes the visual field of a person in a 3D environment and the Inter-Relation Pattern Matrix (IRPM) as a tool for evaluation.

For the tracking part, Hybrid Joint-Separable (HJS) filter is used. The tracker gives the position of the head and feet of each person. Computational result based evaluation is done with the other counterparts in terms of accuracy/efficiency.

**GIZ detection (2015) [30].** This method detects groups based on proxemics. Group Interaction Energy (GIE) feature, Attraction and Repulsion Features (ARF), Granger Causality Test (GCT), and Additional Features (AF) are proposed in this method. Tests are also conducted by combining these features. This method allows people to be connected loosely. The evaluation is done on the basis of computational accuracy and efficiency.

**3D skeleton reconstruction using patch trajectory (2017) [64].** This algorithm works in two stages. First, it takes images from different views as input and produces 3D body skeletal proposals for people using 2D pose detection. Secondly, it refines those using a 3D patch trajectory stream and provides temporally stable 3D skeletons. Authors evaluate the method quantitatively and qualitatively yielding an accuracy of 99%. The limitation of this method lies in its dependency on 2D pose detection and the computation time complexity.

**Learning methods for HPE (head pose estimation) and BPE (Body pose estimation) (2017) [127].** This method uses a jointly learning framework for estimating the head, body orientations of targets, and f-formations for conversational groups. The evaluation matrices used are – HP (head pose) error, BP (body pose) error, and FF (f-formation) F1 score. This method is compared with Inter-Relation Pattern Matrix (IRPM), Interacting Group Discovery(IGD), Hough transforms-based (HVFF), Graph Cut (GC), and Game-Theoretic methods, and the results are not very different in percentage accuracy.

**Method using GAMUT (Group based Meta-classifier learning using local neighborhood Training) (2018) [50].** This method aims at estimating the f-formation of each pair of people in a group as a pairwise relationship in the scene. This method works in two steps: Prepossessing and GAMUT. In the prepossessing stage, raw tri-axial acceleration signals are converted to pairwise feature representations and these are used as samples in GAMUT. In the GAMUT stage, the same size of the local neighborhood is used per window size. The results are computationally evaluated in terms of accuracy and efficiency of detection.

**Method based on pedestrian motion estimation (2018) [105].** This method works in three parts – ego-centric pedestrian detection, pedestrian motion estimation, and group detection using joint motion and proximity estimation. The pedestrian detectors result in bounding boxes (BB) with two features – a position of pedestrians and size of the BB. An optical feature is used for motion estimation. Then, joint pedestrian proximity and motion estimation are used for detecting groups while considering the depth data. The evaluation is done in terms of real-life human experience study using robots and humans.

**Method based on Multi-Person 2D pose estimation (2019) [93].** This method firstly estimates the position of the human skeletal characteristic points in the image plane and calculates the Euclidean distance between those points. Then Part Affinity Fields (PAFs) feature is applied to find the distance based on the Euclidean distance. A curve-fitting approach is used for validation purposes. No need for prior information about camera parameters/features of the scene is needed. The evaluation is performed on the basis of human experience study in real-life scenes using robots and humans.

**Bagged tree (2019) [57].** The proposed algorithm works in three steps – data-set deconstruction, pairwise classification, and reconstruction. Authors evaluate this algorithm with three ML classification models – weighted KNN, bagged trees, and logistic regression, where bagged tree model achieve better result in pairwise accuracy, precision, recall, and F1-score. But this method still needs to be trained on larger datasets and also on richer features. The evaluation is done with human experience study with robots.

**RoboGEM (2019) [124].** RoboGEM is an unsupervised algorithm that detects groups from an ego-centric view. This method works using three main modules – pedestrian detection module "P", pedestrian motion estimation module "V ", and group detection module "G". In the first module, an off-the-shelf pedestrian detector (YOLO) is used that provides bounding boxes for each person in the image. In the second module, V is estimated using optical flow. In the last module, the human group detection is performed using joint motion and proximity estimation. The authors compared this method with existing approaches using Intersection-over-Union (IoU), false positives per image (FPPI), and depth threshold matrices. The evaluation is done with human experience with robots.

Table 5 lists down the surveyed papers on the basis of rule-based and learning-based AI approaches. From the algorithmic trends, it is evident that learning-based approaches are slightly more predominant in recent years. However, both the methods are equally explored over the years and in recent pasts. Learning-based methods tend to be more accurate than their rule-based counterparts. Examples of such methods are: [60], [81], [64], [105], [57], and [124]. Table 6 lists down the accuracy and efficiency of the different approaches. This signifies the detection quality of the methods and techniques. From the survey, it can be established that mostly unsatisfactory accuracy can be seen in rule-based approaches more than learning-based models. And, as expected the main reason in general for low accuracy lies in the inability of the methods to detect dynamic groups as well as multiple groups in the scene accurately. But one interesting observation is that accuracy is largely impacted by the camera vision as well. Basically, low accuracy can be seen in exo-vision input methods and datasets. Similarly, real-timeliness is another issue with methods dealing with dynamic and multiple groups. There is no prominent impact of camera views, datasets, and method types in this case. Readers may refer to the online Appendix which contains the detailed comparison of methods and techniques under rule-based static AI approaches in Table 1 and machine learning-based approaches in Table 2.

Table 5. Classification based on approach/method for group and f-formation detection.

| Classification | References |
|---|---|
| Classical Rule based AI methods [Fixed model based learning and prediction method based on certain geometric assumptions and/or reasoning (Section 6.1).] | Approach behavior [112], sociologically principled method [37], proposed model [117], The Compensation (or Equilibrium) Model, The Reciprocity Model, The Attraction-Mediation Model, The Attraction-Transformation Model [85], rapid ethnography method [79], digital ethnography [91], GroupTogether system [78], museum guide robot system [143], extended f-formation system [46], Multi-scale Hough voting approach [115], HFF (Hough for f-Formations), DSFF (Dominant-sets for f-Formations), [114], PolySocial Reality (PoSR)-F-FORMATION [41], two-Gaussian mixture model,O-space model [52], Wifi based tracking, Laser-based tracking, Vision-based tracking [51], heat map based f-formation representation [44], [128], group tracking and behavior recognition [48], search-based method [45], Estimating positions and the orientations of lower bodies) [142], Kendon's diagramming practice [90], GROUP [131], Graph-Cuts for f-formation (GCFF) [116], [68], head, body pose estimation (HBPE) [121], Link Method, Interpersonal Synchrony Method [104], Frustum of attention modeling [129], f-formation as dominant set model [145], HRI motion planning system [141], footing behavior models (Spatial-Reorientation Model, Eye-Gaze Model) [97], MC-HBPE(matrix completion for head and body pose estimation) method [6], [109], Haar cascade face detector algorithm [73], Haar cascade face detector algorithm [94], [130], Approaching method [111], Measuring Workload Method [80], [95], [96], [103], f-formation as dominant set model [146] |
| Machine Learning based AI methods [Data-driven models for learning and prediction using Supervised, Semi-supervised, Unsupervised and Reinforcement learning (ML/DL) or any such techniques (Section 6.2).] | [12],IR tracking techniques [54], SVM classifier [144], GRID WORLD SCENARIO [86], graph-based clustering method [60], [38], [29], Hidden Markov Model(HMM) [49], proposed method with o-space and without o-sapce(SVM) [108], Region-based approach with level set method [71], IRPM(Inter-Relation Pattern Matrix) [23], graph-based clustering algorithm [126], voting based approach [77], Hidden Markov Models (HMMs) [81], SVM [31], Transfer Learning approaches [102], method with Hidden Markov Model( [59], head pose estimation technique [11], [9], Matrix Completion for Head and Body Pose Estimation (MC-HBPE) [8], GIZ detection [30], [7], Supervised Correlation Clustering (CC) through Structured Learning [119], Long-Short Term Memory (LSTM), Hough-Voting (HVFF) [2], Long Short Term Memory (LSTM) [3], 3D skeleton reconstruction using patch trajectory [64], Human aware motion planner [28], [35], Learning Methods for HPE and BPE [127], GAMUT(Group bAsed Meta-classifier learning Using local neighborhood Training) [50], Group detection method [105], [57], Long Short Term Memory (LSTM) network [107], RoboGEM (Robot-Centric Group Estimation Model) [124], Multi-Person 2D pose estimation [93], Staged Social Behavior Learning (SSBL) [47], multi-class SVM classifier [18] |
| Other studies | Wizard-of-Oz [61], Wizard-of-Oz paradigm [15], Wizard-of-Oz [75] |

Table 6. Comprehensive list of AI methods (Rule based and Learning based) for group/interaction/f-formation detection and their Detection Quality: Accuracy (Excellent/Very good/Good/Average/Poor) and Efficiency (Real-time/Near real-time/ Non real-time) comparison. "Pink" color cells highlight dissatisfying accuracy and "Red" color cells highlight real-timeness issues. *Full forms: SVM (Support Vector Machine), HMM (Hidden Markov Model), HFF (Hough for f-formations), DSFF (Dominant-sets for f-formations), GCFF (Graph-Cuts for f-formation), HBPE (Head and body pose estimation), CC (Correlation Clustering), LSTM (Long-Short Term Memory), RNN (Recurrent Neutral Network) LBFGS (Limited memory Broydon-Fletcher-Goldfarb-Shanno), SGD (Stochastic Gradient Descent), MC-HBPE (Matrix Completion for HBPE), GAMUT (Group bAsed Meta-classifier learning Using local neighbourhood Training), SSBL (Staged Social Behavior Learning), CRF (Conditional Random Field), DT (Decision Tree).*

| **Method** | **Accuracy** | **Efficiency** |
|---|---|---|
| Classical Rule based AI methods (Section 6.1) | | |
| Approach behavior (2009) [112] (SVM) | average | real time |
| Sociologically principled method (2011) [37] (DT) | good (precision 75%) | real time |
| A proposed model (2011) [117] (DT) | good | real time |
| Museum guide robotic system (2012) [143] | very good | real time |
| Extended f-formation system with temporal encoded IS (2013) [46] (DT) | excellent | real time |
| Multi-scale Hough voting approach (2013) [115] (DT) | good | real time |
| HFF (Hough for f-formations) (2013) [114] | performed better using people's position and orientation | — |
| DSFF (Dominant-sets for f-formations) (2013) [114] | performed better when only position information is available | — |
| Wifi based tracking, Laser-based tracking, Vision-based tracking (2013) [51] | good | — |
| Heat map based f-formation representation (2013) [44] (DT) | (Temporal en- coded IS gives a precision of 99.9% and IS gives 80%) | — |
| Estimating positions and the orientations of lower bodies (2014) [142] | (face detection works for one face but Human regions and upper parts can be extracted from a range of images) | — |
| (2014) [128] | very good (most over 80% and for one dataset 100%) | real time |
| Group tracking and behavior recognition (2014) [48] | good | real time |
| Search based method (2014) [45] | good | — |
| Euclidean distance-based NN classifier, ARCO (array-of-co-variance) head pose classifier, WD (weighted) classifier, Multi-view SVM [102] | — | — |
| Graph-Cuts for f-formation (GCFF) (2015) [116] | very good (0.97, 0.94, 0.84, 0.85, 0.92, 0.89) | real time |
| Head and body pose estimation (HBPE) (2015) [121] (DT) | good (0.79 , 0.82 for different dataset ) | real time |
| GROUP (2015) [131] (DT) | excellent (93%) | real time |
| Link Method (2016) [104] | — | real time (2.5 ms) |
| Interpersonal Synchrony Method (2016) [104] | — | real time (10ms) |
| F-formation as dominant set model (2016) [145] | good (71% F-measure) | real time |
| Matrix completion for head and body pose estimation (MC-HBPE) method (2017) [6] | average (49.8°HP 51.6°BP error) | real time |
| Haar cascade face detector algorithm (2017) [73] (DT) | good | real time |
| Haar cascade face detector algorithm (2017) [94] (DT) | good | real time |
| Meta-classifier [50] | — | — |
| Measuring Workload Method (2018) [80] | — | — |
| F-formation as dominant set model (2018) [146] | good (71% F-measure) | real time |
| (2019) [95] (DT) | — | — |
| Machine Learning based AI methods (Section 6.2) | | |
| IR tracking technique with SVM (2010) [54] | good (77.81%) | real time |
| Extraversion, neuroticism (SVM) (2010) [144] | (66%, 75%) | real time |
| Graph-based clustering method (2011) [60] | excellent (all singletons precision 95%, limitation: only for standing) | real time |
| Continued on next page | | |
| Region-based approach with level set method (2012) [71] | average in real environment, good in simulated environment | — |
| Method with o-space and without o-space (SVM) (2012) [108] | good (based on casual, normal, and abnormal behaviour) | real time |
| Hidden Markov Model(HMM) (2012) [49] (Haar-feature classifier, 3-layer artifical neural network (ANN)) | good (66%) | real time |
| Three-dimensional IRPM (Inter-Relation Pattern Matrix) (2013) [23] | very good (89%) (11% false) | non real time (people need to be in same interaction for at least 10 second) |
| Voting based approach (2013) [77] (SVM) | very good | real time |
| Hidden Markov Models (HMMs) (2013)[81] | 0.56 (physical), 0.72 (psychophysical) | real time |
| Graph-based clustering algorithm (2013) [126] (SVM classifier) | very good | real time |
| Poselet detection model (SVM) (2014) [31] | poor | — |
| (2014) [11] (SVM) | very good | real time |
| Method with Hidden Markov Model (2014) [59] | good (64%) | near real time |
| Transfer Learning approach [102] (Euclidean distance-based NN classifier, ARCO head pose classifier, WD classifier, Multi-view SVM) | — | — |
| Matrix Completion for Head and Body Pose Estimation (MC-HBPE) [8] (2015) (TSVM) | average | real-time |
| GIZ detection (2015) [30] (SVM) | very good (can detect even on loosely formed condition) | real-time |
| Model with loss function Pairwise and Mitre (2015) [119] (SVM) | both very good | real time |
| Long-Short Term Memory (LSTM) with LBFGS and SGD, HVFF (2015) [2] (RNN) | very good (82% , 73% , 80%) | real time |
| (2015) [9] (SVM) | good (around 70, 80) | real time |
| Matrix based batch learning for Long Short Term Memory (LSTM) with Limited memory BFGS (L-BFGS) and Stochastic Gradient Descent (SGD) (2016) [3] | SGD performs slightly better than LBFGS (78%) good | real time |
| Method uses HMM (Hidden Markov Model) model and SVM combined) (2017) [35] | very good (85.56) | real-time |
| Learning Methods for HPE (Head-pose estimation) and BPE (Body pose estimation) (2017) [127] | very good (0.87,0.84,0.66 on different dataset) | real time |
| Human aware motion planner (2017) [28] | — | — |
| 3D skeleton reconstruction using patch trajectory (2017) [64] | excellent (99 percentage) | real time |
| GAMUT (Group based Meta-classifier learning Using local neighborhood Training) (2018) [50] | very good | near real time |
| Group detection method (2018) [105] | — | — |
| Multi-Person 2D pose estimation (2019) [93] | very good | real-time |
| Long Short Term Memory (LSTM) network/recurrent neural network [107] | confusion matrix (0.92) | near real-time |
| Staged Social Behavior Learning (SSBL) (2019) [47] | — | non real time |
| (2019) [57] | — | real time |
| RoboGEM (2019) (SVM) [124] | excellent | real time |
| Skeletal key point detection (2020) (SVM, CRF) [18] | excellent | real time |

## 7 CATEGORIZATION OF DETECTION CAPABILITIES AND SCALE

This section puts forward a categorization of the surveyed literature on the basis of group/interaction detection capabilities and scale for a method. After surveying the literature and the methods, it is evident that detection of groups in scenes is a non-trivial task and many factors are to be considered in the process as well. In real-life scenes, there can be both static groups of people interacting without much movement and there can also be groups with constant movement. There can be cases like group members leaving a group or new members joining a group. Group dynamics also is an important factor to be considered. People may sway or move their bodies occasionally too. Apart from these, methods also need to consider a single group or multiple existing groups in a scene. Outliers to one group can be a part of another group or can be noise at the global level. Fig. 11 depicts a taxonomy of group detection in interaction scenarios in real-life cases.

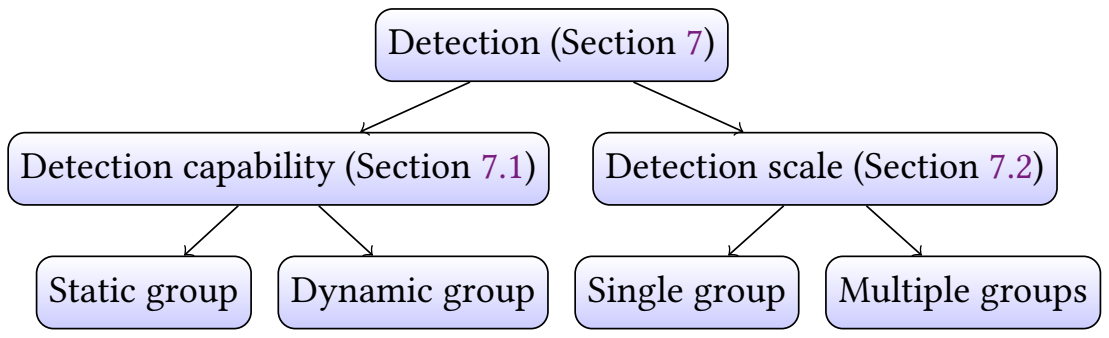

Fig. 11. Taxonomy for Detection capability and scale for groups/interactions and formations.

### 7.1 Detection capability

Methods need to attend to both static and dynamic groups in interactions and formations. Here, we do categorization of this aspect.

**Static group scene.** A static scene means people in the scene are not moving. The people interacting in a group or formation do not change groups or new people do not join a group while interaction is in progress. The people within the group do not sway or change head/body pose and orientation that can affect f-formation detection. In such cases, it is easier to detect groups and formations. No temporal aspect in the scene is to be considered and the method can work on a single image. In [31] and [93], single image is used from a single egocentric camera for detection. Mostly indoor scenes like conferences, group discussions, coffee breaks, and meetings, such static groups can be found.

**Dynamic group scene.** In the case of a dynamic scene, people tend to move in groups, also referred group dynamics. New people can join a group and/or existing people can leave a group. Also, some people participating in an interaction may temporarily change their head/body pose and orientation a bit; this necessarily does not mean that the formation has changed. In such cases, it becomes very difficult for an algorithm to detect the group or formation in the interaction scenario. As a result, the methods need to consider the temporal information of the scene utilizing a sequence of images over a window. A sequence of image/image stream is taken over a particular period of time. In [38], the video data is used which is 10 frames per second for detecting dynamic groups and interactions. In [29], the surveillance videos are used for experiment purposes. Similarly, [115] utilizes video feed from a cocktail party [133] for its experiments, which is one frame in 5 seconds. Fig. 12 depicts the dynamic scene and group scenario. The EGO-GROUP dataset [4] has a video of an indoor laboratory setup. The video consists of 395 image frames. The specialty of this video is that the people in the scene are not static in one position and they change position/orientation and location with time. On the right-hand side of the figure, we put forward four instances of the image sequence where four different types of groups/interactions and formations are visible for the same four people in the scene. This type of dynamism should be handled by the detection methods with efficiency considering temporal aspects of the scenes. In outdoor scenes such as waiting rooms, stations, airports, restaurants, theatres, and lobbies, dynamic groups are mostly encountered. Table 7 summarizes the references into two detection capability types found in the literature.

### 7.2 Detection scale

Since we need different methods depending on how many groups are there in a captured image or video, the detection scale plays an important role.

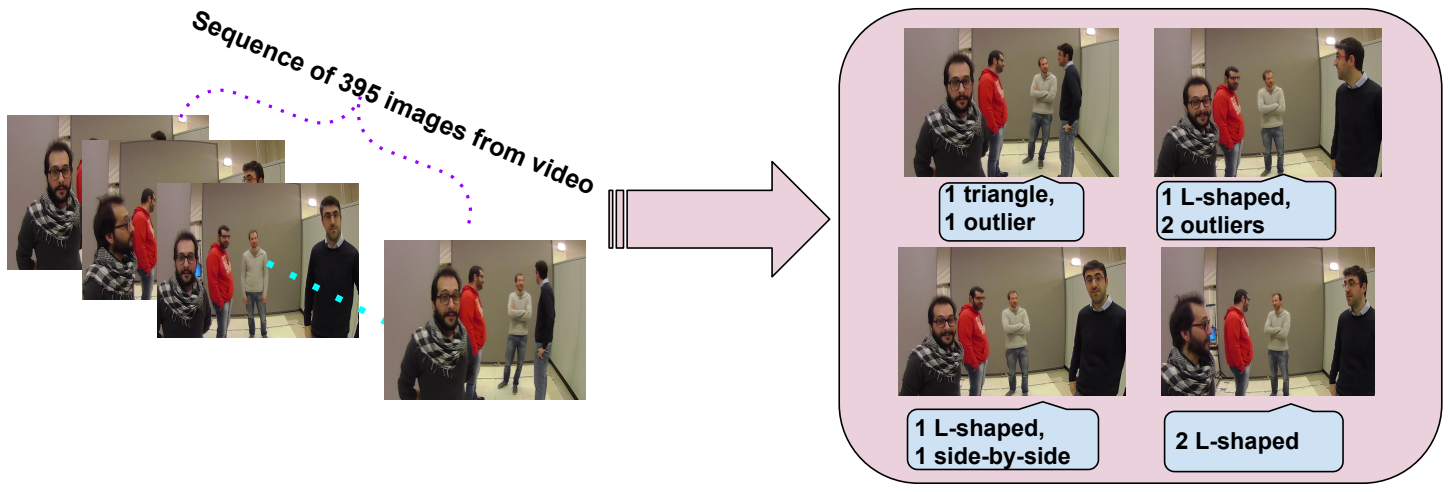

Fig. 12. Dynamic behaviour of groups/people in a video/image sequence from EGO-GROUP [4].

Table 7. Classification based on group/interaction and formation detection capability.

| **Classification** | **References** |
|---|---|
| Static scene detection | [93], [50], [92],[109], [145], [129], [104], [116], [44], [114], [100], [128], [31], [102], [69], [47] |
| Dynamic scene detection | [124], [58], [103], [107], [105], [50], [111], [92], [80], [40], [130], [16], [64], [6], [141], [35], [94], [127], [73], [125], [62], [113], [104], [3], [9], [8], [90], [68], [116], [121] , [118], [7], [131], [119], [2], [54], [75], [144], [60], [82], [85], [117], [38], [79], [61], [12], [29], [91], [78], [71], [143], [84], [108], [49], [77], [115], [95], [51], [126], [23], [46], [41], [128], [11], [59], [48], [142], [45], [67], [112], [66], [140], [30], [15], [37], [18], [146] |

**Single group detection.** When a sensor/camera detects only one interacting group in the scene, the work is easily done. The stream of images sequence can have multiple groups as well. But all the methods do not have the capability to detect multiple groups simultaneously. In some cases, single group detection is useful when a robot needs to detect a single group of interest in a scene or environment and join the group for interaction/discussion. The datasets used for this kind of detection are mostly captured indoor (for example office and panoptic studio [33]) or outdoor (mostly private datasets). Other publicly available datasets are BEHAVE [89] and YouTube videos which can also be used for such purposes. In the case of ego-view camera-based detection methods, single group detection is the primary focus.

**Multiple group detection.** When there is more than one interacting group or formation in a scene, the detection methods need special attention. Sometimes there is only one interacting group in the scene along with some additional people who are not actively involved in the interaction. Those cases can also be considered under the same umbrella and are quite challenging too. This kind of detection is useful for finding how many groups are there or finding a particular group in a diverse scene in surveillance/monitoring applications. There exists some datasets comprising of such scenarios – coffee break dataset [36], EGO-GROUP [4], SALSA [134], cocktail party [133], GDet [19], Synthetic [43], Idiap Poster Data [63], and FriendsMeet2 (GM2) [20, 22]. Beyond these, some researchers have used their own (private) datasets. In [38] and [115], the authors experimented with such datasets where there is more than one group in the scene (the party data). Similarly, in [29], the surveillance videos are used as data where there can be more than one group in the captured video. Table 8 classifies the literature on the basis of group/interaction detection scale. The multiple group detection scenario normally comes in exo-view based methods. Fig. 14 depicts three scenarios from a renowned dataset, EGO-GROUP [4]. Fig. 14a shows a single triangular formation with one outlier in an indoor environment. Fig. 14b depicts a viz-a-viz formation in an outdoor situation. Fig. 14c shows two groups, one triangular and one L-shaped formation in an indoor situation.

Table 9 comprehensively lists all the surveyed datasets in the literature. A total of 70 datasets have been mentioned out of which only 14 are publicly available, 51 are private to the authors/researchers and 5 are not known. 15 of the datasets have outdoor scenes (mostly from public area captures), 41 have indoor scenes, and only 5 of them have both types of scenes. 36 out of the list have multiple group scenarios in the scenes and 19 have single group scenes and 15 are not known. Ego-vision scenes of groups and interactions are seen in 20 datasets, whereas 38 datasets have exo-vision or global view images of groups, 1 dataset has both of them, and camera-view is not known for 11 datasets. Fig. 13 gives a comprehensive idea about the taxonomy of datasets (training/testing) generally used in group/interaction and formation detection tasks.

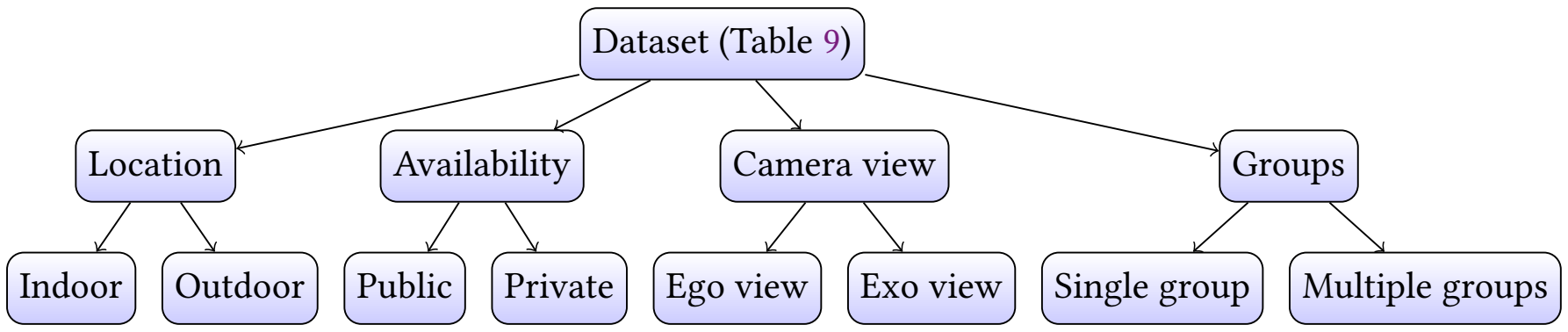

Fig. 13. Taxonomy for datasets surveyed for groups/interactions and formation detection.

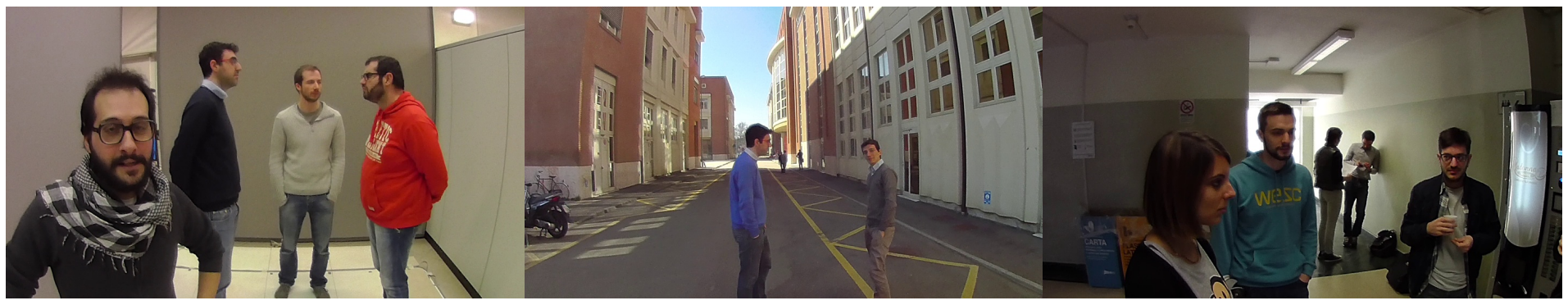

(a) Single group with outlier (laboratory) (b) Single group (outdoor) (c) Multiple groups (cafeteria)

Fig. 14. Images from EGO-GROUP [4] dataset depicting indoor and outdoor scenes with single and multiple group interactions.

Table 8. Classification based on group/interaction and formation detection scale.

| Classification | References |
|---|---|
| Single group detection | [58], [47], [57], [111], [92], [80], [40], [97], [130], [109], [16], [64], [35], [94], [73], [125], [62], [113], [3], [90], [68], [116], [118], [131], [2], [75], [144], [82], [88], [85], [86], [117], [79], [61], [12], [91], [78], [143], [84], [108], [49], [44], [74], [46], [41], [100], [42], [65], [45], [69], [67], [112], [66], [140], [15], [95], [18] |
| Multi-group detection | [124],[93], [96], [103], [107], [105], [50], [6], [127], [145], [141], [129], [104], [9], [8], [116], [121], [7], [119], [54], [60], [29], [71], [114], [77], [115], [126], [23], [81], [128], [11], [59], [31], [102], [48], [142], [30], [38], [92], [37], [146] |

Table 9. Comprehensive list of datasets (training/testing) and their types, used in group/interaction and formation detection methods surveyed in the literature. The color coding is done for the readers to perceive the data easily.

| **Dataset** | **View (Ego/Exo)** | **Single/Multiple group (s)** | **Indoor/outdoor [area]** | **Availability (Public/Private)** |
|---|---|---|---|---|
| TUD Stadtmitte [13] | Ego | Multi-gp | outdoor [public] | private |
| HumanEva II [13] | Ego | Multi-gp | indoor | private |
| SALSA [134] | Exo | Multi-gp | indoor | public |
| BEHAVE database [89] | Exo | Multi-gp | outdoor [public] | public |
| TUD Multiview Pedestrians [13] | Exo | Multi-gp | outdoor [public] | private |

Continued on next page

| | | | | |
|---|---|---|---|---|
| CHILL [29] | Exo | Multi-gp | — | — |
| Benfold [24] | — | — | — | — |
| MetroStation [29] | Exo | Multi-gp | indoor [public] | private |
| TownCentre [25] | Exo | Multi-gp | outdoor [public] | private |
| Indoor [29] | Exo | Multi-gp | indoor | private |
| SI (Social Interactions) [83] | Exo | Multi-gp | outdoor [public] | public |
| Coffee-room scenario [23] | Exo | Multi-gp | indoor | private |
| CoffeeBreak [36] | Exo | Multi-gp | outdoor [private] | public |
| Collective Activity [14] | Ego | Multi-gp | outdoor/indoor | private |
| PETS 2007 (S07 dataset) [23] | Exo | Multi-gp | indoor [public] | private |
| Structured Group Dataset (SGD) [132] | Exo | Multi-gp | indoor/outdoor [public] | public |
| EGO-GROUP [4] | Ego | Multi-gp | indoor/outdoor | public |
| EGO-HPE [5] | Ego | Multi-gp | indoor/outdoor | public |
| Mingling [59] | Exo | Multi-gp | indoor | private |
| MatchNMingle [27] | Exo | Multi-gp | indoor | public |
| CLEAR [120] | Exo | Single-gp | indoor | private |
| Greece [102] | Exo | Multi-gp | indoor | private |
| DPOSE [101] | Exo | Multi-gp | indoor | private |
| BIWI Walking Pedestrians [98] | Exo | Multi-gp | outdoor [public] | private |
| Crowds-By-Examples (CBE) [76] | Exo | Multi-gp | outdoor [public] | private |
| Vittorio Emanuele II Gallery (VEIIG) [17] | Exo | Multi-gp | indoor [public] | private |
| UoL-3D Social Interaction [34] | Ego | Single-gp | indoor | public |
| Cocktail Party [133] | Exo | Multi-gp | indoor | public |
| Social Interaction [33] | — | Single-gp | indoor | public |
| GDet [19] | Two monocular cameras, located on opposite angles of a room | — | indoor | public |
| Idiap Poster Data (IPD) [63] | Exo | Multi-gp | outdoor | public |
| Classroom Interaction Database [77] | Exo | Multi-gp | indoor | private |
| Caltech Resident-Intruder Mouse [26] | — | — | — | — |
| UT-Interaction [110] | Exo | Multi-gp | outdoor | private |
| PosterData [60] | Exo | Multi-gp | outdoor | private |
| Friends Meet [20, 22] | Exo | Multi-gp | outdoor | public |
| Discovering Groups of People in Images (DGPI) [32] | Exo | Multi-gp | indoor | private |
| Prima head pose image [53] | Ego | Single-gp | indoor | private |
| NUS-HGA [30] | — | Single-gp | indoor | private |
| [54] | Exo | Single-gp | indoor | private |
| [144] | Exo | Single-gp | indoor | private |
| [60] | Exo | Multi-gp | indoor | private |
| [38] | Exo | Multi-gp | outdoor [public] | private |
| [86] | Exo | — | — | — |
| [49] | Ego | Single-gp | indoor | private |
| Dataset using Narrative camera [87] | Ego | Single-gp | indoor | private |
| [3] | Ego | Single-gp | indoor/outdoor [public] | private |

Continued on next page

| | | | | |
|---|---|---|---|---|
| [50] | Exo | Multi-gp | indoor | private |
| [105] | Ego | Multi-gp | outdoor | private |
| Laboratory-based dataset containing distance measures at three key distances, one laboratory-based dataset with distance measures from three predefined distances, dataset with distance measurements collected in a crowded open space [93] | – | – | – | private |
| RGB-D pedestrian dataset [124] | Ego | Multi-gp | outdoor [public] | private |
| [61] | – | – | indoor | private |
| [112] | – | – | indoor | private |
| [117] | – | – | indoor | private |
| [79] | – | – | indoor | private |
| [85] | Ego | Single-gp | indoor | private |
| Youtube videos [91], [90] | – | – | indoor | private |
| [78] | Exo | Single-gp | indoor | private |
| [46] | Exo | – | indoor | private |
| In shopping mall [51] | Ego | – | – | private |
| [44] | Exo | Single-gp | indoor | private |
| [21] | Ego | – | – | private |
| [15] | Ego | Single-gp | indoor | private |
| DGPI dataet[129] | – | – | – | – |
| [141] | Exo | Single-gp | indoor | private |
| [109] | Ego, Exo | Single-gp | indoor | private |
| [130] | Ego | Single-gp | indoor | private |
| [73] | Ego | Single-gp | indoor | private |
| [111] | Ego | Single-gp | indoor | private |
| [80] | Ego | – | – | private |

## 8 CATEGORIZATION OF EVALUATION METHODS

The most important part of the formation or interaction detection framework (Fig. 4) is the evaluation methodologies. The conventional methods to compare methods and techniques in such vision tasks are accuracy and efficiency. The accuracy defines how accurately a method detects/predicts or recognizes an f-formation. The efficiency parameter relates to the real-timeliness aspect of the method. Apart from these the papers in the surveyed literature also speak about simulation-based evaluation and human experience study-based evaluations (for robotic applications specifically). Fig. 15 shows the simple taxonomy of evaluation methods for various group/interaction and formation detection methods or algorithms.

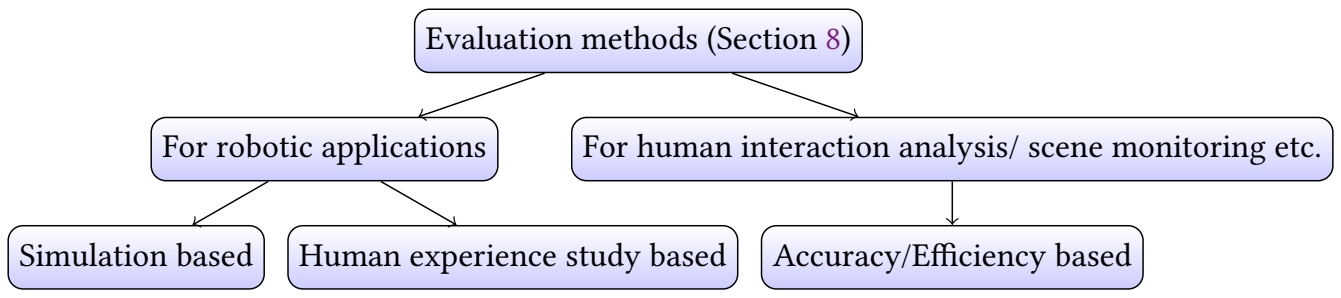

Fig. 15. Taxonomy for evaluation methodology for group/interaction and f-formation detection.

**Simulation based evaluation.** This type of evaluation is conducted using simulation tools. The simulators have different features and have the ability to simulate the real world in complex environments. A range of simulators are used in the surveyed literature – Gazebo [96], RoboDK [106], and Webot [39]. Nowadays researchers are also focused on using Virtual Reality (VR) or Augmented Reality (AR) technologies for evaluation purposes. The evaluations are performed mainly to access the perception of a virtual robot or an autonomous agent. The question to be answered is, how well a simulated robot (in a simulated environment) can perceive a group of simulated people involved in an interaction. Secondly, after detection, is the simulated robot joining the group naturally without discomforting the simulated people (see Section 2.3 and Fig. 3). Parameters like stopping distance for the robot, orientation, and pose based on the perceived group pose/orientation and angle of approach depending on the group's angle and position should be considered. Extensive discussion on these factors post group/interaction detection by a robot or autonomous agent is out of the scope of this survey.

**Human experience study based evaluation.** This type of evaluation is based on testing the detection methods using ego-vision robots or on a real scenario with human participants as evaluators. A questionnaire is provided to the human participators to rate the quality of the method being used by the robot in real scenarios. The questions and parameters similar to simulation-based evaluation can be considered in this case as well but with a real robot perceiving human groups (who are also the evaluators) and interaction. In real-life scenarios, the groups are not static and tend to move when a member joins and leaves the group. Accordingly, the robot or the autonomous agent must detect the changes in group formation, orientation, and pose to re-adjust itself in a natural and more human-like manner without causing any comfort to other humans.

**Accuracy/Efficiency evaluation without using robot or simulators.** This kind of evaluation is based on the accuracy or efficiency aspects of the methods but not tested in the real environment or by using robots. Here, the focus is mainly to evaluate the computational aspects of the methods/algorithms without evaluating the usability in real-life applications like robotics. However, applications like human behavior/interaction analysis, scene monitoring, and surveillance depend entirely on such evaluation. Table 10 classifies the surveyed papers on the basis of the evaluation strategy adopted. It also shows the descriptions/names of simulators in the simulation-based category.

Table 10. Classification based on evaluation methods and strategy.

| **Classification** | **References** |
|---|---|
| Simulation based evaluation (robotic simulators/virtual environment) | 2D grid environment simulated in Greenfoot [86], simulated the process of deformation of contours using P-spaces represented by Contours of the Level Set Method [71], [45], Robot Operating System (ROS) implementation of PedSim [104], a simulated avatar embodied confederate [97], Gazebo [96], a simulator using Unity 3D game engine [47] |
| Human experience study based evaluation (with real robots) | [12], [61], [75], [82], [85], [117], [143], [49], [51], [126], [81], [15], [141], [94], [28], [109], [73], [130], [111], [105], [80], [93], [95], [57], [124], [18] |
| Accuracy/Efficiency evaluation (without robot, only computation) | [54], [144], [60], [37], [38], [79], [29], [91], [78], [71], [108], [46], [115], [114], [23], [77], [44], [45], [128], [11], [59], [31], [102], [48], [142], [45], [8], [90], [116], [121], [30], [131], [129], [104], [3], [145], [35], [127], [6], [64], [50], [107], [103], [146] |

## 9 APPLICATION AREAS

Group or interaction detection has seen vast applications in many areas of computer vision. Specifically speaking, with the emergence of robotics and AI, this domain has realized its true potential. In this paper, we categorize the application landscape into two broad areas: robotic applications and other vision applications. Further, these have been broken down into five groups as summarized in Table 11. The robot vision implies the applications where the robot's camera is placed in an ego-centric view for finding the groups only, but there is no purpose of initiation of interaction with a human. In Human-robot interaction, f-formation detection is used to detect the group in order to participate in the interaction with fellow human beings autonomously. In telepresence, a remote person uses the robot to interact with a group of people. In such a scenario, the semi-autonomous robot can detect the group and join them while the remote human operator can control the robot to adjust its positioning.

Scene monitoring is useful for analyzing indoor or outdoor scenes with people interacting and forming groups and f-formations for various activities. On the other hand, human behavior and interaction analysis refer to the behavior between humans and how they are interacting based on the situation. Furthermore, visual analytics in big data has empowered the domain beyond imagination. People are trying to use these technologies in various aspects of life. In the current scenario of the Covid-19 pandemic, we can utilize this technology in monitoring social distancing in human groups and interactions as well. As already mentioned, telepresence robotics can be utilized by doctors/nurses and other medical staff to attend to patients in remote locations without physically being present.

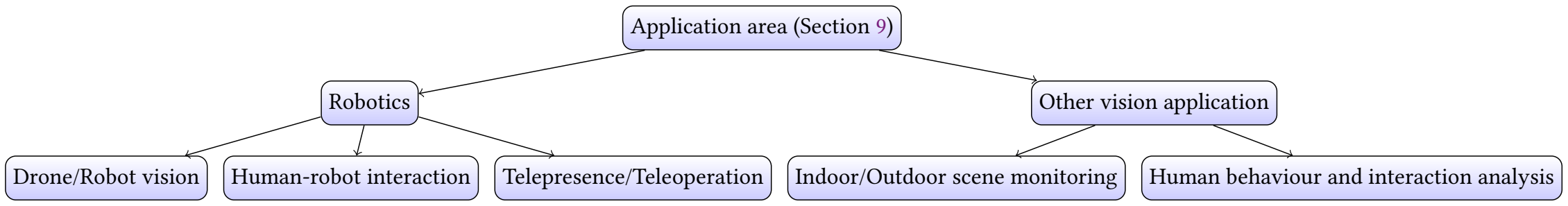

Fig. 16. Taxonomy for application areas for group/interaction and f-formation detection.

Table 11. Classification based on targeted application areas.

| Application area | References |
|---|---|
| Drone/Robotic vision | [58], [64], [6], [127], [125], [145], [129], [113], [104], [8], [116], [121], [131], [119], [2], [75], [60], [29], [78], [114], [77], [115], [23], [46], [41], [100], [128], [11], [31], [102], [48], [123], [30], [37], [18] , [146] |
| Human-robot interaction | [124], [93], [47], [57], [105], [50], [111], [92], [80], [130], [109], [16], [141], [35], [62], [68], [118], [75], [82], [88], [85], [86], [117], [61], [12], [143], [84], [49], [74], [51], [81], [52], [65], [142], [69], [67], [112], [66], [140], [15] |
| Telepresence/Teleoperation technologies | [96], [92], [94], [73] |
| Indoor/outdoor Scene monitoring and surveillance | [58], [144], [97], [75], [82], [51], [99] |
| Human behaviour and interaction analysis | [103], [107], [40], [127], [104], [3], [9], [90], [7], [54], [144], [82], [38], [79], [61], [91], [71], [108], [44], [126], [42], [59], [45], [123] |
| Covid-19 and social distancing | Scope of future research |

## 10 LIMITATIONS, CHALLENGES AND FUTURE DIRECTIONS: A DISCUSSION

The survey is treated based on a generic framework of concern areas about group/interaction detection using the theory of f-formation (see Section 3 and Fig. 4). It addressess various identified modules and concern areas such as camera view and availability of other sensor data, datasets, feature selection, methods/techniques, detection capabilities/scale, evaluation methodologies, and application areas.

- The existing methods have almost equal share of fixed rule-based (Fig. 9) and learning-based (Fig. 10) approaches (Tables 5 and 6; In online Appendix Tables 1 and 2). Researchers need to orient their research towards data-driven approaches using deep learning and reinforcement learning paradigms for handling complex situations. Meta-learning can also be explored on large-scale combined datasets. The complex scenarios in detection tasks can be easily solved using big data and visual analytics [99]. Apart from that, representing data in the form of a graph can solve many performance issues in terms of accuracy and efficiency. The graph neural networks (GNN) such as graph convolution networks (GCNN) can also be a potential candidate to create appropriate models. A combination of recurrent neural networks (RNN), convolution neural networks (CNN), and/or graph recurrent networks (GRNN) can also be explored for identifying more accurate and promising detection models.

- The problems like dynamism in groups (people leaving/joining the group dynamically or changing position and orientation within the group) and occlusions of people pose serious challenges and limitations to the current state-of-the-art methods in terms of accuracy and efficiency. Researchers can think about devising rules based on reasoning and geometry to detect application-specific groups and interactions. A combination of rules, geometry-based reasoning along data-driven models can also be explored to improve detection quality. Apart from detecting the group and formation alone, methods should be designed to detect the orientation and pose of the group itself (see [18]). This can facilitate a good approach direction and angle (natural and human-like) for robots to join the group.
- The major challenge with the datasets is their availability. Creating good quality (large scale) vision datasets (for training and testing) is a mammoth task in itself but has its own research/academic merit. The only 20% of the surveyed datasets are publicly available (Table 9). The researchers can publish more of their privately created datasets as benchmarks for people to experiment. Another limitation is the availability of public ego-vision datasets with only 30% of the total being ego-vision. Researchers can think of creating more first-person view datasets by merging/fusing existing datasets in a meaningful manner. As for the exo-vision datasets, accuracy is low, so the researchers can look into this direction by creating more robust datasets for global view scenes. Indoor datasets dominate the scenario currently with merely 29% of the datasets being outdoor. Researchers need to create more outdoor datasets for the sake of applications pertaining to surveillance and outdoor scene monitoring. Finally, the researches have limited their methods and datasets to only a few major formations like face-to-face, triangular/circular, side-by-side, and L-shaped. No literature or state-of-the-art methods speak about dealing with a comprehensive list of formations (as explained in Section 2.2). Researchers should concentrate on devising methods and creating datasets to solve these limitations.
- Detection capabilities need attention with respect to dynamic scenes (Fig. 12) as well as multiple groups (Fig. 14). The literature is rich in taking care of most of the aspects of detection (Tables 7 and 8). However, some more research attention is required in cases of occlusion, background clutter, and lighting conditions. Researchers can use reinforcement learning and deep learning models for these problems. Also, appropriate datasets need to be prepared at a larger scale.
- Evaluation of the methods remains a challenge in the current literature (Table 10). Mostly, computational evaluation has been performed in terms of accuracy and efficiency (Table 6). But in a problem like a group/interaction detection, human experience studies and/or simulation-based studies are important to establish the effectiveness of the method in various applications like robotics, telepresence, and social surveillance (see Section 8, Fig. 15 and Table 11). The researchers need to orient their studies in this respect as well. Apart from that, most of the methods yield good accuracy but achieving real-time solutions maintaining good accuracy is a concern. The explorers can think of designing lightweight models for real-time detection of groups and interactions for dynamic scenes.
- Feature selection/extraction plays a major role in any computer vision problem, and group detection is no exception. The existing literature lacks discussion about the use of proper features and the selection of proper approaches to extract useful and differentiating features. Apart from the visual features, researchers can also think about non-visual features such as audio or speech as future research trends. There is also a possibility of temporal feature selection and extraction for dynamic groups.

- Applications of this domain can be widely seen in robotics, surveillance, human behavior analysis, and telepresence technologies (Table 11 and Fig. 16). However, we can also think about using this technology in Covid-19 related applications such as monitoring of social distancing norms and others.
- We also have discussed about two types of camera views: Ego and Exo-views (Fig. 6 and Table 4). Ego vision is used predominantly for robotics related applications. Methods using ego-view cameras for input are less compared to exo-view cameras. The main reason behind this roots down to the scarcity of public ego-view datasets for training the models. Researchers can also direct their research on designing detection models which can be created on a hybrid system of camera views and sensors. The visual as well as other forms of inputs combined can be used for better detection and prediction tasks. Various combinations of camera views and positions can be experimented with for better scene capture and robust dataset creation for learning models.

Fig.s 17 and 18 summarizes the limitations, challenges and future directions/opportunities for all the concern areas of this survey framework.

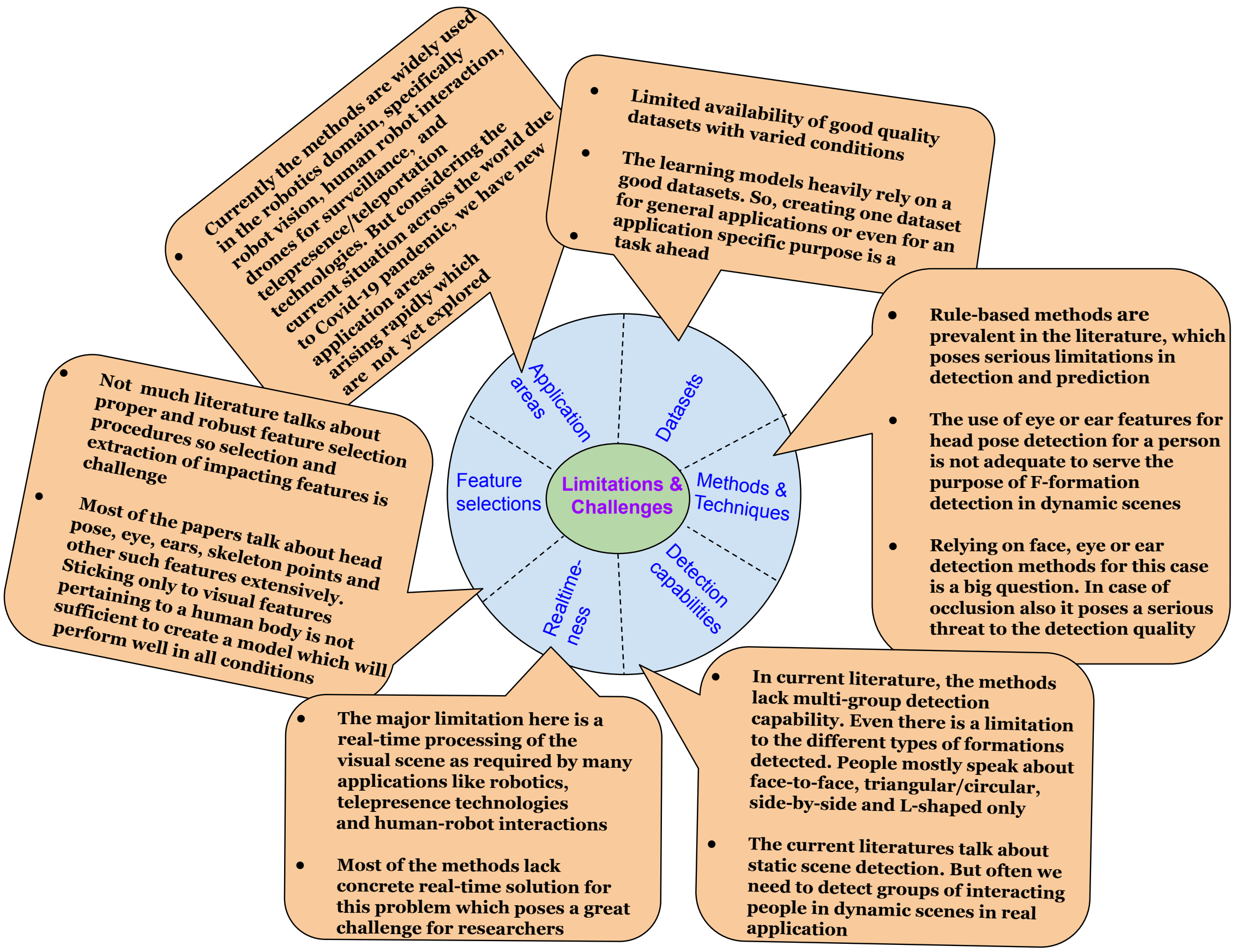

Fig. 17. Limitations and challenge in the various concern areas of our survey framework (see Section 3).

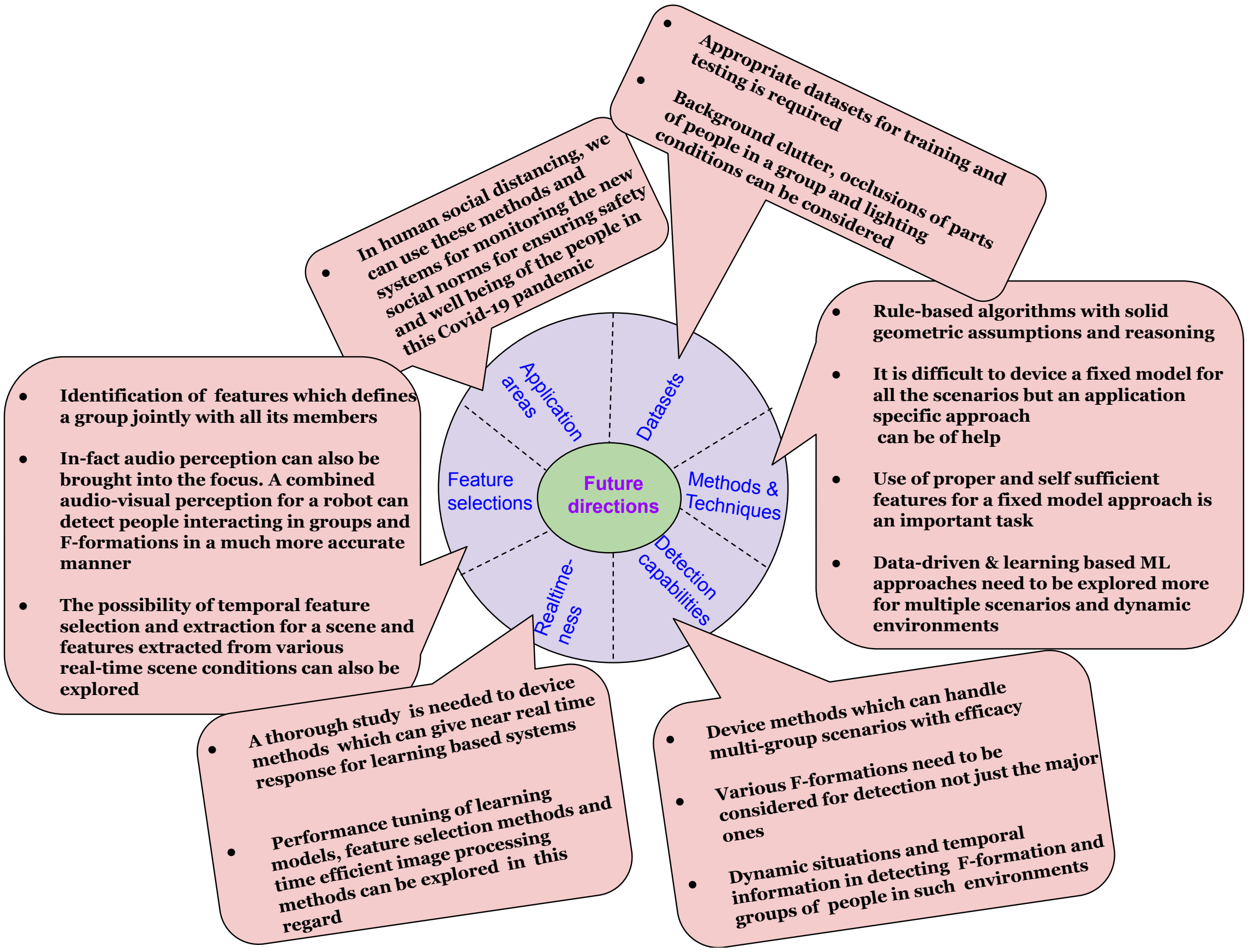

Fig. 18. Future research directions/opportunities in the various concern areas of our survey framework (see Section 3).

## 11 CONCLUSIONS

With the emergence of computer vision, robotics, multimedia analytics, etc., the world is changing for good with the progress of artificial intelligence. Computation systems and autonomous agents are expected to show more human-like behavior and capability. One of the most important problems in this domain persists to be group/interaction detection and prediction using f-formation. Although some research has been conducted in the last decade, much more progress is still envisioned. This survey aims at generalizing the problem of group/interaction detection via a framework, which is referred to as the theme of this survey as well. This article presents a comprehensive glance at all the concern areas of this defined framework. This includes definitions of various f-formations, input camera views and sensors, datasets, feature selection, algorithms, detection capability and scale, quality of detection, evaluation methodologies, and application areas. The article also discusses the limitations, challenges, and future scope of research in this domain. The researchers can try and solve some of the unattended problems in this domain with the help of more recent and efficient approaches in deep learning, reinforcement learning, and meta-learning paradigms. For example, the family of Graph neural networks (GNN, GCNN, GRNN, etc.) remains a very potent tool to solve the problems in this domain. A combination of different neural networks can also be explored to create efficient models in application-specific cases.

# Online Appendix to: Detecting Socially Interacting Groups using F-formation: A Survey of Taxonomy, Methods, Applications, Challenges and Future Research Directions

HRISHAV BAKUL BARUA, Robotics & Autonomous Systems (Cognitive Robotics), TCS Research, India
THEINT HAYTHI MG, Myanmar Institute of Information Technology, Myanmar
PRADIP PRAMANICK, Robotics & Autonomous Systems (Cognitive Robotics), TCS Research, India
CHAYAN SARKAR, Robotics & Autonomous Systems (Cognitive Robotics), TCS Research, India

The Appendix consists of two long tables as a part of electronic supplementary material. These Tables 1 and 2 summarizes the surveyed methods and techniques for group, interaction and F-formation detection on the basis of various parameters. The Table 1 compares among the static rule based AI methods, while Table 2 informs about the Machine learning or Deep learning based techniques. The tables consists of some short forms in the table headers used to label each column and in other rows: **ML** (Machine Learning), **DL** (Deep Learning), **RL** (Reinforcement Learning), **Su** (Supervised), **UnS** (Unsupervised), **SS** (Semi-supervised), **S** (Single group), **M** (Multiple groups), **Exo** (Exocentric vision or global view), and **Ego** (Egocentric or first person view).

Table 1. Comparison between Static AI based methods for detecting F-formation and social interactions.

| Methods and Techniques | Static AI approach | | | Detection capability | | Features selected | Datasets used for testing | Camera and/or Sensors |
|---|---|---|---|---|---|---|---|---|
| | Fixed rules | Assumptions | Geometric rea-son-ing | Static (S/M) | Dynamic (S/M) | | | |
| Wizard-of-Oz (2006) [51] | ✗ | ✗ | ✓ | ✗ | ✓(S) | Follow, show, validate | indoor [51] | ActivMedia Performance PeopleBot |
| approach behavior (2009) [88](SVM for classification) | ✗ | ✓ | ✗ | ✗ | ✓ | shapes of trajectory, velocity, and direction | indoor | Robovie |
| Wizard-of-Oz approach(2010) [58] | ✓ | ✗ | ✗ | ✗ | ✓(S) | head, torque, whole body condition | indoor | GestureMan-4 robot |

Continued on next page

Authors' addresses: Hrishav Bakul Barua, Robotics & Autonomous Systems (Cognitive Robotics), TCS Research, Kolkata, India, hrishav.barua@tcs.com; Theint Haythi Mg, Myanmar Institute of Information Technology, Mandalay, Myanmar, theinthaythimg@gmail.com; Pradip Pramanick, Robotics & Autonomous Systems (Cognitive Robotics), TCS Research, Kolkata, India, pradip.pramanick@tcs.com; Chayan Sarkar, Robotics & Autonomous Systems (Cognitive Robotics), TCS Research, Kolkata, India, sarkar.chayan@tcs.com.

| | | | | | | | | |
|---|---|---|---|---|---|---|---|---|
| sociologically principled method(2011) [34] (DT) | ✗ | ✓ | ✗ | ✗ | ✓(M) | Sampling, Voting, O-space validation | GDet [17], CofFeeBreak [33] | Exo |
| proposed model (2011) [92] (DT) | ✗ | ✓ | ✗ | ✗ | ✓(S) | gaze zone, sight zone and front zone | indoor [92] | humanoid robot |
| rapid ethnography method (2011) [62] | | | | ✗ | ✓(S) | environmental features | indoor [62] | notes and sketches |
| The Compensation (or Equilibrium) Model, The Reciprocity Model, The Attraction-Mediation Model, The Attraction-Transformation Model (2011) [66] | | | | ✗ | ✓(S) | likeability and gaze behavior | indoor | robot |
| digital ethnography(2012) [71] | | | | ✗ | ✓(S) | content map | youtube videos | |
| GroupTogether sensing system (2012) [61] | ✗ | ✓ | ✗ | ✗ | ✓(S) | interactions using a combination of motion sensors, radio modules with coarse-grained range finding capability, and overhead Kinect-based depth camera tracking | indoor [61] | Exo, sensors |
| museum guide robot system (2012) [108] | ✓ | ✗ | ✗ | ✗ | ✓(M) | distance between the robot and the visitors(DRV), distance between the robot and the exhibits(DRE), FOV of the robot(RFOV), face direction | indoor [108] | humanoid robot Robovie-R Ver.3 (Vstone) |
| extended F-formation system(2013) [39] (DT) | ✗ | ✓ | ✗ | ✗ | ✓(S) | heat map based feature | indoor [39] | Kinetic depth sensors, Exo cameras |
| Multi-scale Hough voting approach(2013) [90] (DT) | ✓ | ✗ | ✗ | ✗ | ✓(M) | Fixed-cardinality group detection, group merging | Synthetic Data, CoffeeBreak Seq1, CoffeeBreak Seq2 [33], CocktailParty[109] | Exo |
| HFF (Hough for F-Formations)(2013) [89] | ✓ | ✗ | ✗ | ✗ | ✓(M) | position, head orientaion | Idiap Poster Data (IPD)[89], Coffee Break (CB) Dataset[33] | Exo |
| DSFF (Dominant-sets for F-Formations)(2013) [89] | ✓ | ✗ | ✗ | ✗ | ✓(M) | affinity matrix | Idiap Poster Data (IPD)[89], Coffee Break (CB) Dataset[33] | Exo |
| two-Gaussian mixture model(2013) [45] | ✓ | ✗ | ✗ | | | human heading | | |
| O-space model(2013) [45] | ✗ | ✓ | ✗ | | | position and orientation | | |
| Wifi based tracking, Laser-based tracking, Vision-based tracking , position tracking system(2013) [44] | ✓ | ✗ | ✗ | ✗ | ✓(S) | | in shopping mall [44] | Service robots |

Continued on next page

| | | | | | | | | |
|---|---|---|---|---|---|---|---|---|
| heat map based F-formation (2013) [37] (DT) | ✗ | ✗ | ✓ | ✗ | ✓(S) | social attributes(human face, etc) | indoor [37] | sensor, Exo, Web |
| Estimating positions and the orientations of lower bodies) (2014) [107] | | | | ✓(M) | ✗ | head position together with facial orientation | [18] | Ego |
| (2014) [99] | | | | ✓(M) | ✓(M) | positions of the persons and head/body orientations, 2D histogram | PosterData [50], CocktailParty[109], CoffeeBreak[34], Synth[34], GDet[34] | Exo |
| group tracking and behavior recognition(2014) [41] | | | | ✗ | ✓(M) | the average of the intra-object distance, the average standard deviations of speed and direction | recorded in subway | Exo |
| Search-based method(2014)[38] | ✗ | ✓ | ✗ | ✓(S) | ✗ | 2D Local Coordinate System (LCS) | indoor [38] | -Ego wearable sensors |
| (2015) [70] | | | | ✗ | ✓(S) | time stamped map | Youtube videos | Exo, Ego |
| (2015) [14] | | | | ✓(S,M) | ✗ | | indoor [14] | Ego |
| Graph-Cuts for F-formation (GCFF) (2015) [91] | | | | ✓(M) | | | Synthetic [36], Coffee Break [33], IDIAP Poster Data[52], Cocktail Party [104], GDet[17] | Exo |
| head,body pose estimation (HBPE) and FFE accuracy(2015) [95] (DT) | ✗ | ✓ | ✗ | ✗ | ✓(M) | HOG of head and body | Coffee Break [33], Cocktail Party [104] | Exo |
| GRUPO(2015) [102] (DT) | ✗ | ✓ | ✗ | ✗ | ✓(M) | head orientations, lower body orientation | Cocktail Party [104] | Exo |
| Grounded Theory Method(2015) [55] (DT) | | | | ✗ | ✓(M) | human orientations and face reaction | | FROG robot |
| Link Method (2016) algorithm is inspired by learning and forgetting curves combined with proxemics. [81] | | | | ✗ | ✓(M) | low level features | Friends Meet[19], SALSA[105] | Exo |
| Interpersonal Synchrony Method(2016) exploits interpersonal synchrony to refine clusters of people obtained mixing proxemics and the intersections of the 2D fields-of-view of people.[81] | | | | ✗ | ✓(M) | low level features | Friends Meet[19], SALSA[105], synthetic data set | Exo |

Continued on next page

| | | | | | | | | |
|---|---|---|---|---|---|---|---|---|
| Frustum of attention modeling(2016) [100] | | | | ✓(Single frame) | ✓(M)(Multi-frame) | | FriendsMeet2(FM2) [19], iscoveringGroupsofPeopleinImages(DGPI) [28], PosterData [50], CocktailParty [104], CoffeeBreak [33], Synth [34], GDet [17] ,DGPI dataset | Exo |
| F-formation as dominant set model (2016) [110] | ✗ | ✓ | ✗ | ✓(M) | ✗ | orientation invariant features, social pior features, proximity features | Idiap Poster Data [53] | Exo |
| HRI(Human Robot Interaction motion planning system (2017) [106] | ✗ | ✓ | ✗ | ✗ | ✓(M) | omnidirectional mobility feature | indoor[106] | Exo |
| footing behavior models(Spatial-Reorientation Model, Eye-Gaze Model)(2017)[76] (DT) | | | | ✗ | ✓(S) | Spatial probability distributions, Gaze fixation lengths | | VR |
| MC-HBPE method(matrix completion for head and body pose estimation) (2017) [5] | ✗ | ✓ | ✗ | ✗ | ✓(M) | body and head features | SALSA [105] | Exo |
| (2017) [85] | | | | ✓(S) | ✗ | shape, size, distance, angle | indoor [85] | Nao robot |
| (2017) [101] (DT) | | | | ✗ | ✓(S) | Body orientaion and gaze behavior | indoor | UWB(Ultra-wideband) localization beacons, Kinect, Audio sensors |
| Haar cascade face detector algorithm(2017) [57] (DT) | ✗ | ✓ | ✗ | ✗ | ✓(S) | quadrants | indoor | telepresence robot, Ego |
| Haar cascade face detector algorithm (2017) [73] (DT) | ✗ | ✓ | ✗ | ✗ | ✓(S) | quadrants | Prima head pose image dataset [46] | mobile robot telepresence (MRP) |
| Approaching method(2018) [87] (DT) | | | | ✗ | ✓(S) | human-friendly features | indoor [87] | MIRob platform, Ego |
| Measuring Workload Method(2018) [63] | ✗ | ✓ | ✗ | ✗ | ✓(S) | NASA-TLX(NASA Task Load Index) Scores, Dual-task Method Scores | indoor [63] | Ego robot |
| F-formation as dominant set model (2018) [111] | | | | ✓(M) | ✗ | orientation invariant features, social pior feature | Idiap Poster Data[53] , SALSA[105] | Exo |
| (2019) [74] (DT)<br><br>(2019) [75] (DT) | ✗ | ✓ | ✗ | ✗ | ✓ | Robot Positioning Spot (RPS)<br>2D frustum , autonomous features | EGO-GROUP[3] | Egoview robot<br><br>Mobile Robotic Telepresence System in Simulation Environment |

Continued on next page

| | | | | | | | | |
|---|---|---|---|---|---|---|---|---|
| (2019) [80] | | | | ✗ | ✓(M) | speaking turns | MatchNMingle dataset[24] | Simultaneous Speakers |

Table 2. Comparison between ML based methods for detecting F-formation and social interactions.

| **Methods** | **Approach classification** | | | **Learning paradigm** | | | **Detection capability** | | **Feature selection** | **Dataset** | **Camera and/or Sensors** |
|---|---|---|---|---|---|---|---|---|---|---|---|
| | **ML Based** | **DL Based** | **RL Based** | **Su** | **UnS** | **SS** | **Static (S/M)** | **Dynamic (S/M)** | | | |
| (2004) [11] | ✗ | ✓ | ✗ | ✓ | ✗ | ✗ | ✗ | ✓(S) | state diagram | | - Egoview robot |
| IR tracking technique (2010) [47] , (Gaussian Mixture Model, Naive Bayes, Support Vector Machine) | ✓ | ✗ | ✗ | | | | ✗ | ✓ | data points of location and orientation | indoor[47] | - Exo |
| SVM cassifier(2010) [109] | ✓ | ✗ | ✗ | ✓ | ✗ | ✗ | ✗ | ✓(S) | Minimum distance, Velocity, Number of intimate, personal and social relationships, | indoor [109] | - Exo |
| graph-based clustering method (2011)[50] | | | | | | | ✗ | ✓(M) | Dominant set, Affinity matrix, Socially Motivated Estimate of Focus Orientation(SMEFO) | indoor [50] | -Exo |
| Gaussian clustering, Expectation-Maximization (EM) learning method (2011) [35] | ✓ | ✗ | ✗ | ✗ | ✓ | ✗ | ✗ | ✓ | Gaussian components | public area outdoor | - Exo |
| GRID WORLD SCENARIO(three level hierarchical modeling approach)(2011) [67] | | | | | | ✗ | ✗ | ✓(S) | Affinity value | | - Exo (software agent bird's eye view) |
| coupled adaptive classifier learning (2012)[26] | | | | ✗ | ✗ | ✓ | ✗ | ✓(M) | Body and head pose features | TUD Multiview Pedestrians dataset [12], Benfold dataset [21], CHILL dataset [26], MetroStation dataset[26], Indoor dataset[26], TownCentre dataset[22], | - Exo |
| Region-based approach with level set method (2012) [56] | ✗ | ✓ | ✗ | ✓ | ✗ | ✗ | ✓(M) | ✗ | | surveillance camera | - Exo |

Continued on next page

| | | | | | | | | | | | |
|---|---|---|---|---|---|---|---|---|---|---|---|
| proposed method with o-space and without o-sapce(SVM)(2012) [84] | | | | | | | ✗ | ✓ | salient motion features | own dataset:SI (Social Interactions) Dataset[65], YouTube CCTV videos, BEHAVE database[69] | - Exo |
| Hidden Markov Model(HMM)(2012) [42] (Haar-feature classifier, 3-layer artifical neural network (ANN)) | ✓ | ✗ | ✗ | ✓(all data are labeled) | ✗ | ✗ | ✓(using image processing) | ✗ | body posture and head pose estimation(torso and hand positions (excluding the arms), body alignment, head pose (given both as a normal vector and as pitch and yaw angles), as well as the two spatial group arrangement features) | indoor [42] | -Ego |
| three-dimensional IRPM(Inter-Relation Pattern Matrix)(2013)[20] | ✓ | ✗ | ✗ | | | | ✗ | ✓(M) | Subjective View Frustum | coffee-room scenario[20],PETS 2007(S07 dataset) | -Exo |
| voting based approach (2013) [60](SVM) | | | | | | | ✗ | ✓(M) | Histogram of Oriented Gradient (HOG) feature for head pose | Classroom Interaction Database[60], UT-Interaction Dataset [86] , Caltech Resident-Intruder Mouse Dataset [23] | Exo |
| Hidden Markov Models (HMMs)(2013)[64] | | | | | | | ✗ | ✓(M) | Physical feature, psychophysical features(Mehrabian's physical features and Hall's psychophysical features) | indoor [64] | -Using Prime-Sensor |
| graph-based clustering algorithm (2013)[97](SVM classifier) | ✓ | ✗ | ✗ | | | | ✗ | ✓(M) | motion information and local interaction information using Local Group Activity(LGA) descriptor | CoffeeBreak dataset [33], Collective Activity dataset [13] | -Exo |
| Poselet detection model (SVM) (2014)[29] | | | | | | | ✓(M) | ✗ | Individual Pose feature,3D Estimation, INdividual Unary Feature, Pairwise interaction feature, cross-validated confidence values | Structured Group Dataset (SGD) [103] | -Exo |
| Head pose estimation technique(2014)[10](structure SVM) | | | | | | | ✗ | ✓(M) | | EGO-GROUP [3], EGO-HPE [4] | -Ego |

Continued on next page

| | | | | | | | | | | | |
|---|---|---|---|---|---|---|---|---|---|---|---|
| method with Hidden Markov Model(2014)[49] | ✓ | ✗ | ✗ | | | | ✗ | ✓(M) | | mingling dataset | -single worn accelerometer, audio sensing |
| Transfer Learning approaches [79] (2014) (ARCO-Xboost, Euclidian distance-based NN classifier, ARCO head pose classifier, WD classifier, Multi-view SVM) | ✓ | ✗ | ✗ | | | | ✓(M) | ✗ | | CLEAR dataset [94], DPOSE dataset [78] ,Greece dataset | - Exo |
| Matrix Completion for Head and Body Pose Estimation (MC-HBPE)[8] (2015) | ✓ | ✗ | ✗ | | | | ✗ | ✓(M) | head and body visual features | SALSA dataset [7], [105] | - Exo |
| GIZ detection (2015) [27](SVM) | ✓ | ✗ | ✗ | | | | ✗ | ✓(S) | Group Interaction Energy(GIE), Attraction and Repulsion Features(ARF) | BEHAVE dataset[69], NUS-HGA dataset [27] | - Exo |
| [6] | | | | | | | ✗ | ✓(M) | target's states based on ground locations | SALSA dataset [105] | - Exo |
| Supervised Correlation Clustering (CC) through Structured Learning[93](SVM) | | | | | | | ✗ | ✓(M) | physical identity and soical identiy | BIWI Walking Pedestrians dataset [77], Crowds-By-Examples (CBE) dataset [59], Vittorio Emanuele II Gallery (VEIIG) dataset [15] | - Exo |
| Long-Short Term Memory (LSTM) (2015) [1](RNN) | ✗ | ✓ | ✗ | | | | ✓(S) | ✗ | distance and orientation features | dataset using narrative camera [68] | - Ego |
| Hough-Voting (HVFF)[1] | | | | | | | | | | dataset using Narrative camera[68] | -Ego |
| Head pose estimation method) (2015) [9](SVM) | | | | ✓ | ✗ | | ✗ | ✓(M) | facial landmark feature | EGO-GROUP [3] | - Ego |
| matrix based batch learning for Long Short Term Memory (LSTM) with Limited memory BFGS(L-BFGS) and Stochastic Gradient Descent(SGD) (2016) [2] | ✗ | ✓ | ✗ | ✗ | ✓ | ✗ | ✗ | ✓(S) | people localization, face orientation estimation, 3D people localization | indoor and outdoor egocentric dataset | - Ego |
| Continued on next page | | | | | | | | | | | |
| Uses HMM model with two states)(2017) [32](SVM) | | | | | | | ✗ | ✓(S) | Upper Joint Distances,Body relative orientations,Temporal Orientation similarity,O-space based features,QTCC relation,QTCC histogram | UoL-3D Social Interaction Dataset [31] | -Exo |
| Learning Methods for HPE and BPE (2017) [98] | ✓ | ✗ | ✗ | | | | ✗ | ✓(M) | head and body features | Cocktail Party [109][104], Coffee Break [33], SALSA [7][105] | - Exo |
| Human aware motion planner (2017) [25] | ✗ | ✗ | ✓ | | | | ✗ | ✓(M) | Gaussian distribution, metric map | | - |
| 3D skeleton reconstruction using patch trajectory (2017) [54] | | | | | | | ✗ | ✓(S) | node proposals, part proposals, skeletal proposals | Social Interaction Dataset [30] | - Exo |
| GAMUT(Group bAsed Meta-classifier learning Using local neighborhood Training)(2018) [43] | ✓ | ✗ | ✗ | ✓ | ✗ | ✗ | ✓(M) | ✓(M) | pairwise feature | indoor [43] | -Exo |
| Group detection method (2018) [82] | ✗ | ✓ | ✗ | ✗ | ✓ | ✗ | ✗ | ✓(M) | the position of pedestrians and size of the BB(Bounding boxes) | outdoor [82] | - Ego-centric telepresence robot |
| Multi-Person 2D pose estimation (2019) [72] | ✓ | ✗ | ✗ | | | | ✓(M) | ✗ | Part Affnity Fields(PAFs) feature, facial or body features | laboratory-based dataset containing distance measures at three key distances, one laboratory-based dataset with distance measures from three predefined distances, dataset with distance measurements collected in a crowded open space[72] | - Ego |
| Long Short Term Memory (LSTM) network/recurrent neural network [83] | ✗ | ✓ | ✗ | | | | ✗ | ✓(M) | pairwise representations, label extractions | MatchNMingle dataset [24] | -wearable sensors |
| Staged Social Behavior Learning (SSBL) (2019) [40] | ✗ | ✗ | ✓ | | | | ✓(S) | ✗ | feature-map | | -SoftBank Pepper robot |

Continued on next page

| | | | | | | | | | | | |
|---|---|---|---|---|---|---|---|---|---|---|---|
| (2019) [48] | ✓ | ✗ | ✗ | | | | ✗ | ✓(S) | | SALSA dataset [8] | -Egoview robot |
| RoboGEM (Robot-Centric Group Estimation Model) (2019) (SVM)[96] | ✗ | ✓ | ✗ | ✗ | ✓ | ✗ | ✗ | ✓(M) | identifies people, proximity, velocity | own RGB-D pedestrian dataset [96] | -Ego Double Telepresence Robot |
| Skeletal key point detection SVM,CRF)(2020)[16] | ✓ | ✗ | ✗ | | | | ✗ | ✓(S) | | EGO-GROUP [3], own dataset | -Ego |